\documentclass{article} 
\usepackage{iclr2026_conference,times}

\usepackage{amsmath,amsfonts,bm}

\def\eqref#1{equation~\ref{#1}}

\def\1{\bm{1}}

\DeclareMathAlphabet{\mathsfit}{\encodingdefault}{\sfdefault}{m}{sl}
\SetMathAlphabet{\mathsfit}{bold}{\encodingdefault}{\sfdefault}{bx}{n}

\usepackage{hyperref}
\usepackage{url}

\usepackage{graphicx}
\usepackage{booktabs}       
\usepackage{amsmath}
\usepackage{amssymb}
\usepackage{tcolorbox}
\usepackage{xspace}
\usepackage{enumitem}
\usepackage{fontawesome5}
\usepackage{marvosym}
\usepackage[utf8]{inputenc}
\usepackage{array}
\usepackage{tabularx}
\usepackage{dashrule}
\usepackage{arydshln}
\usepackage{tcolorbox}
\usepackage{multirow}
\usepackage{subcaption}
\usepackage{lipsum}
\usepackage{amssymb}
\usepackage{pifont}
\usepackage{titletoc}
\tcbuselibrary{breakable}
\usepackage{xcolor} 

\usepackage{float}
\usepackage{graphicx}       
\usepackage{wrapfig}      
\usepackage{subcaption}     
\usepackage{makecell}
\usepackage{soul}
\usepackage{duckuments}
\usepackage{fix-cm}
\usepackage{listings}
\usepackage{xspace}
\usepackage{tikz}
\usepackage{wrapfig}
\usepackage{tcolorbox}
\usepackage{colortbl}

\usetikzlibrary{tikzmark}
\newcommand{\myfontsize}{\normalsize}

\usepackage[T1]{fontenc}   

\lstdefinestyle{python}{
    language=Python,
    basicstyle=\ttfamily\small,
    keywordstyle=\color{blue}\bfseries,
    commentstyle=\color{green},
    stringstyle=\color{red},
    numberstyle=\tiny\color{gray},
    showstringspaces=false,
    frame=single,
    breaklines=true,
    backgroundcolor=\color{lightgray!20}
}

\definecolor{pythonblue}{RGB}{52, 101, 164}
\definecolor{searchgreen}{RGB}{76, 175, 80}
\definecolor{webblue}{RGB}{33, 150, 243}
\definecolor{imagepurple}{RGB}{156, 39, 176}
\definecolor{terminalorange}{RGB}{255, 152, 0}
\definecolor{sqlred}{RGB}{244, 67, 54}
\definecolor{mcpgray}{RGB}{96, 125, 139}

\makeatother
\newcommand{\mytextbox}[2]{\tikzmarknode[draw=#1,thick,inner sep=2pt]{test}{\myfontsize #2}}

\definecolor{myred}{rgb}{0.7, 0.3, 0.0}
\definecolor{myblue}{HTML}{054488}
\definecolor{mygreen}{HTML}{056b34}
\definecolor{myorange}{HTML}{ff8800}
\definecolor{mypurple}{HTML}{8400ff}
\definecolor{mypink}{HTML}{f7acb9}

\newcommand{\red}[1]{\mytextbox{myred}{\textbf{\textcolor{myred}{#1}}}}
\newcommand{\blue}[1]{\mytextbox{myblue}{\textbf{\textcolor{myblue}{#1}}}}
\newcommand{\green}[1]{\mytextbox{mygreen}{\textbf{\textcolor{mygreen}{#1}}}}

\newcommand{\purple}[1]{\mytextbox{mypurple}{\textbf{\textcolor{mypurple}{#1}}}}

\definecolor{codegray}{RGB}{240,240,240}
\definecolor{commentgreen}{RGB}{0,128,0}
\definecolor{keywordblue}{RGB}{0,0,180}
\definecolor{stringred}{RGB}{180,0,0}

\lstdefinestyle{python}{
    language=Python,
    backgroundcolor=\color{codegray},
    basicstyle=\ttfamily\footnotesize,
    keywordstyle=\color{keywordblue}\bfseries,
    stringstyle=\color{stringred},
    commentstyle=\color{commentgreen}\itshape,
    showstringspaces=false,
    breaklines=true,
    frame=single,
    framerule=0pt,
    rulecolor=\color{codegray},
}

\lstdefinestyle{sql}{
    language=SQL,
    backgroundcolor=\color{codegray},
    basicstyle=\ttfamily\footnotesize,
    keywordstyle=\color{blue!60!black}\bfseries,
    stringstyle=\color{stringred},
    commentstyle=\color{gray}\itshape,
    showstringspaces=false,
    breaklines=true,
    frame=single,
    framerule=0pt,
    rulecolor=\color{codegray},
}

\definecolor{mydarkblue}{rgb}{0,0.08,0.45}
\definecolor{mydarkgreen}{RGB}{0, 139, 69}
\definecolor{MAEblue}{RGB}{47 112 182}
\definecolor{SDEblue}{RGB}{28 58 88}
\hypersetup{
	colorlinks=true,
	urlcolor=magenta,
	citecolor=MAEblue,
        linkcolor=blue,
}
\definecolor{mycyan}{cmyk}{.3,0,0,0}
\definecolor{cc1}{rgb}{1.0, 0.44, 0.37}
\definecolor{cc2}{rgb}{0.0, 0.2, 0.6}
\definecolor{cc3}{RGB}{255, 191, 0}
\definecolor{cc4}{RGB}{0, 128, 128}

\newcommand{\ours}{\textsc{VerlTool}}
\newcommand{\ourscolored}{\textbf{{\color{cc2}\textsc{Verl}}{\color{cc1}\textsc{Tool}}}}

\def\adl@drawiv#1#2#3{%
        \hskip.5\tabcolsep
        \xleaders#3{#2.5\@tempdimb #1{1}#2.5\@tempdimb}%
                #2\z@ plus1fil minus1fil\relax
        \hskip.5\tabcolsep}
\newcommand{\cdashlinelr}[1]{%
  \noalign{\vskip\aboverulesep
           \global\let\@dashdrawstore\adl@draw
           \global\let\adl@draw\adl@drawiv}
  \cdashline{#1}
  \noalign{\global\let\adl@draw\@dashdrawstore
           \vskip\belowrulesep}}
\makeatother

\title{VerlTool: Towards Holistic Agentic\\Reinforcement Learning with Tool Use}

\author{
$^{\clubsuit}$Dongfu Jiang$^{*\dagger}$, $^{\diamondsuit}$Yi Lu$^*$, $^\varsigma$Zhuofeng Li$^*$, $^{\clubsuit}$Zhiheng Lyu\thanks{Equal Contribution. Dongfu Jiang is the project leader.} , $^\varrho$Ping Nie, $^{\vartheta, \theta}$Haozhe Wang,\\
\textbf{
$^{\clubsuit}$Alex Su, $^{\varepsilon}$Hui Chen, $^{\varphi}$Kai Zou, $^{\spadesuit}$Chao Du, $^{\spadesuit}$Tianyu Pang, $^{\clubsuit}$Wenhu Chen
}\thanks{Corresponding Author.}\\
$^{\clubsuit}$University of Waterloo, $^{\spadesuit}$Sea AI Lab, 
$^{\diamondsuit}$University of Toronto,  $^\varsigma$Shanghai University, \\
$^{\vartheta}$HKUST, 
$^{\theta}$\includegraphics[height=.7em]{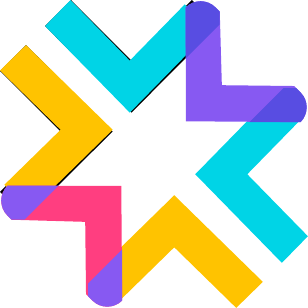}\,M-A-P, 
$^{\varepsilon}$National University of Singapore, $^{\varphi}$NetMind.AI, $^\varrho$Independent \\
\texttt{\{dongfu.jiang@uwaterloo.ca, wenhuchen@uwaterloo.ca\}} \\
}

\iclrfinalcopy 
\begin{document}

\maketitle

\begin{abstract}
Reinforcement Learning with Verifiable Rewards (RLVR) has demonstrated success in enhancing LLM reasoning capabilities, but remains limited to single-turn interactions without tool integration. While recent \textbf{A}gentic \textbf{R}einforcement \textbf{L}earning with \textbf{T}ool use (ARLT) approaches have emerged to address multi-turn tool interactions, existing works develop task-specific codebases that suffer from fragmentation, synchronous execution bottlenecks, and limited extensibility across domains. These inefficiencies hinder broader community adoption and algorithmic innovation. We introduce \ourscolored{}, a unified and modular framework that addresses these limitations through systematic design principles. \ourscolored{} provides four key contributions: \textbf{(1)} upstream alignment with VeRL ensuring compatibility and simplified maintenance, \textbf{(2)} unified tool management via standardized APIs supporting diverse modalities including code execution, search, SQL databases, and vision processing, \textbf{(3)} asynchronous rollout execution achieving near 2$\times$ speedup by eliminating synchronization bottlenecks, and \textbf{(4)} comprehensive evaluation demonstrating competitive performance across 6 ARLT domains. Our framework formalizes ARLT as multi-turn trajectories with multi-modal observation tokens (text/image/video), extending beyond single-turn RLVR paradigms. We train and evaluate models on mathematical reasoning, knowledge QA, SQL generation, visual reasoning, web search, and software engineering tasks, achieving results comparable to specialized systems while providing unified training infrastructure. The modular plugin architecture enables rapid tool integration requiring only lightweight Python definitions, significantly reducing development overhead and providing a scalable foundation for tool-augmented RL research. Our code is open-sourced at \url{https://github.com/TIGER-AI-Lab/verl-tool}.
\end{abstract}

\vspace{-1em}

\begin{figure}[!h]
    \centering
    \includegraphics[width=1.0\linewidth]{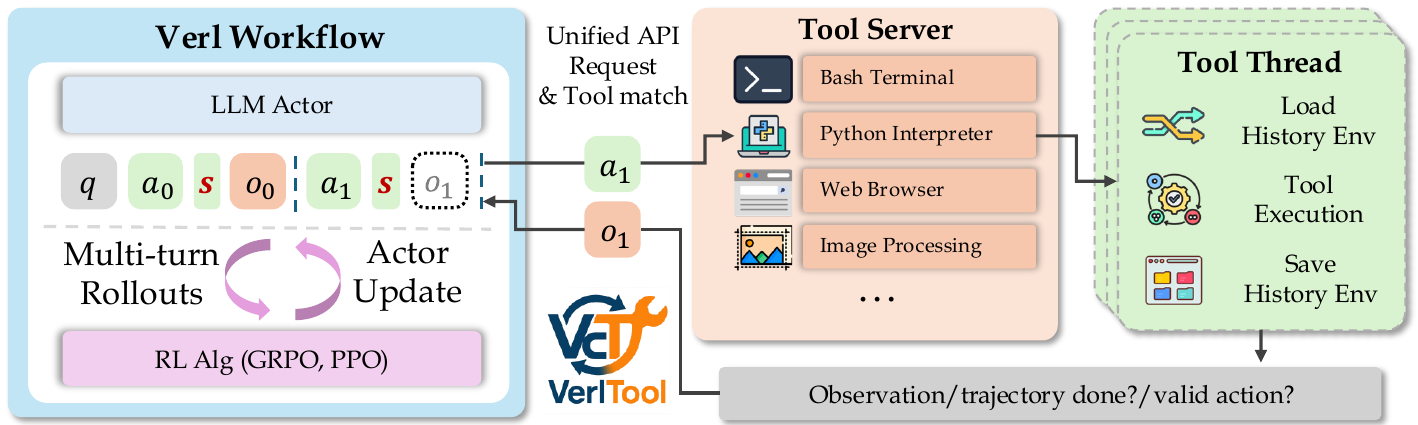}
    \caption{Overview of the \ourscolored{}, a modularized and efficient framework for the \textbf{A}gentic \textbf{R}einforcement \textbf{L}earning with \textbf{T}ool Use (ARLT) training paradigm, where the RL workflow and tool execution are fully disaggregated for both efficiency and extensibility.}
    \label{fig:overview}
\end{figure}

\section{Introduction}
\label{sec:intro}

\begin{quote}
\emph{``We shape our tools, and thereafter our tools shape us.''} \hfill --- Marshall McLuhan
\end{quote}

Large language models (LLMs) such as OpenAI's O-series~\citep{o1} and \textsc{DeepSeek-R1}~\citep{r1} have recently achieved striking advances, surpassing top human performers in challenging domains like mathematics (\textsc{AIME}) and programming (\textsc{LiveCodeBench}~\citep{jain2024livecodebench}, \textsc{Codeforces}~\citep{Quan2025CodeEloBC}). A central driver of this progress is the paradigm of \emph{reinforcement learning with verifiable rewards} (RLVR), which strengthens long-context reasoning during training. Through RLVR, LLMs exhibit emergent cognitive behaviors such as reflection, backtracking, and multi-step reasoning.

Yet these systems remain constrained in a fundamental way: they are unable to interact with the external world. Current LLM reasoning unfolds in a closed, single-turn setting without environmental feedback, often leading to brittle behaviors such as overthinking~\citep{Chen2024DoNT} or hallucination~\citep{Yao2025AreRM}. Conceptually, these models resemble a ``brain in a vat'', locked into self-contained simulations without grounding in interactive or physical reality.

To overcome this isolation, a parallel line of work has explored augmenting LLMs with the ability to use tools. Systems such as \textsc{ToolFormer}~\citep{Schick2023ToolformerLM} and \textsc{OpenHands}~\citep{Wang2024OpenHandsAO} extend models with supervised training on synthetic tool-use data, enabling practical interaction with code interpreters, search engines, or APIs. However, these approaches primarily rely on imitation learning. They lack the agentic autonomy needed to learn directly from feedback and refine their behavior adaptively in open-ended environments.

Recent research begins to bridge this gap by combining tool use with RLVR, giving rise to a new paradigm we term \textbf{ARLT}—\textbf{A}gentic \textbf{R}einforcement \textbf{L}earning with \textbf{T}ool use. In ARLT, LLMs can actively engage with external tools such as code execution environments~\citep{Li2025ToRLST}, search engines~\citep{searchr1}, image manipulators~\citep{pixelreasoner}, and domain-specific APIs~\citep{Feng2025ReToolRL}. This interaction transforms training into a multi-turn, feedback-rich process that not only improves efficiency and reduces token usage but also fosters more robust agentic behaviors.

However, enabling ARLT poses significant challenges from a systems perspective. First, \emph{rollout efficiency} becomes critical: multi-tool trajectories unfold asynchronously, with different tools producing results at varying speeds, demanding scalable asynchronous execution. Second, \emph{tool management} remains fragmented: existing ARLT codebases are often tailored to specific tools, making it difficult to extend or reproduce results. Finally, \emph{multimodal support} is still underdeveloped: while most RL frameworks focus narrowly on text, emerging multimodal reasoning agents (e.g., \textsc{Pixel-Reasoner}~\citep{pixelreasoner}) require handling tool outputs that include images, videos, or other structured modalities in a unified design.

These barriers have slowed community progress, limiting reproducibility, extensibility, and algorithmic innovation. To address them, we introduce \ours{}: an open-source, user-friendly, and efficient framework built on top of \textsc{VeRL}~\citep{sheng2024hybridflow}, designed explicitly for ARLT that supports both text and multimodal training. Unlike prior systems, \ours{} enables multi-turn, stateful agentic training with tool use through four key contributions:
\begin{itemize}[leftmargin=1cm]
\setlength{\itemsep}{0pt}
\setlength{\parskip}{0pt}
    \item \textbf{Upstream Alignment.} \ours{} inherits \textsc{VeRL} as a submodule, ensuring compatibility with upstream updates. This modular separation between RL training and agentic interaction simplifies maintenance and accelerates framework evolution.
    \item \textbf{Unified Tool Management.} We introduce a dedicated tool server with standardized interaction APIs, supporting diverse tools such as code execution, search, SQL/tabular reasoning, and vision utilities. Adding a new tool requires only a lightweight Python definition file, streamlining extensibility for both training and evaluation.
    \item \textbf{Asynchronous Rollouts.} By interacting with tool servers on a trajectory-by-trajectory basis rather than synchronously batch by batch, \ours{} eliminates idle waiting time. This design yields over $2\times$ speedup during rollout execution.
    \item \textbf{Diverse ARLT Tasks.} We have implemented and tested \ours{} on six ARLT tasks, including Math, Search Retrieval, SQL, Visual Reasoning, Web Browsing, and SWE-Bench, achieving competitive performance with previous baselines while trained in a unified framework. We also present common findings in the agentic RL setting across these tasks.
\end{itemize}

In summary, \ours{} provides a principled, extensible, and efficient framework for ARLT, bridging the gap between isolated LLM reasoning and interactive agentic intelligence. By combining upstream-aligned RL infrastructure, unified tool integration, asynchronous execution, and diverse tasks, it paves the way for scalable research and practical deployment of LLMs as tool-using agents.

\section{Related Work} 
\label{sec:related}

\subsection{Reinforcement Learning for Agentic Tool Use}
\label{subsec:arlt}

The integration of reinforcement learning with tool use has emerged as a powerful paradigm for developing adaptive LLM agents. Early tool-calling approaches relied on prompt-based orchestration~\citep{Yao2022ReActSR,Lu2023ChameleonPC,Shen2023HuggingGPTSA}, building on Chain-of-Thought reasoning~\citep{Wei2022ChainOT} and multi-agent frameworks for training-free tool invocation. While instruction-tuned models~\citep{Schick2023ToolformerLM,Kong2023TPTUv2BT,gou2023tora} learned structured calling patterns through supervised learning, they remained largely static and limited to single-turn interactions.

Recent work has demonstrated the advantages of reinforcement learning for tool use, enabling models to optimize their tool-calling policies based on execution outcomes and environmental feedback~\citep{Li2025ToRLST,Feng2025ReToolRL,Moshkov2025AIMO2WS, wang2025code}. This paradigm, which we refer to as \emph{Agentic Reinforcement Learning with Tool use (ARLT)}, extends beyond single-turn verification to support long-horizon, multi-turn interactions. Key characteristics of ARLT include: (1) credit assignment across sequential tool calls, (2) explicit handling of observation tokens from tool responses, and (3) integration with robust, failure-aware execution environments~\citep{Plaat2025AgenticLL,Ke2025ASO}.

This shift from static instruction-following to dynamic, feedback-driven learning has shown effectiveness across diverse domains, including mathematical reasoning with code execution, information retrieval, natural language to SQL generation, and visual reasoning tasks. These applications require agents to probe environments iteratively, adapt to tool feedback, and refine their strategies—capabilities that are difficult to achieve through purely supervised approaches.

\subsection{Agentic RL Training Frameworks}
\label{subsec:agentic_rl_frameworks}

\begin{table}[!ht]
\centering
\small
\caption{Tool support comparison across different frameworks (up to update until August 23, 2025). RAGEN and ROLL focus on the puzzle environments like bandit, which we did not list here.}
\label{tab:tool_supports}
\begin{tabular}{l|cccccc}
\toprule
\textbf{Framework} & \textbf{\textcolor{searchgreen}{\faSearch}} & \textbf{\textcolor{pythonblue}{\faPython}} & \textbf{\textcolor{webblue}{\faGlobe}} & \textbf{\textcolor{terminalorange}{\faTerminal}} & \textbf{\textcolor{sqlred}{\faDatabase}} & \textbf{\textcolor{imagepurple}{\faImage}} \\
& \textbf{FAISS} & \textbf{Python} & \textbf{Web} & \textbf{Bash} & \textbf{SQL} & \textbf{Image} \\
& \textbf{Search} & \textbf{Executor} & \textbf{Search} & \textbf{Terminal} & \textbf{Executor} & \textbf{Processing} \\
\midrule
\textsc{OpenRLHF}~\citep{hu2024openrlhf} & \textcolor{searchgreen}{\checkmark} & \textcolor{searchgreen}{\checkmark} & \textcolor{red}{$\times$} & \textcolor{red}{$\times$} & \textcolor{red}{$\times$} & \textcolor{red}{$\times$} \\
\textsc{VeRL}~\citep{sheng2024hybridflow} & \textcolor{searchgreen}{\checkmark} & \textcolor{searchgreen}{\checkmark} & \textcolor{red}{$\times$} & \textcolor{red}{$\times$} & \textcolor{red}{$\times$} & \textcolor{red}{$\times$} \\
\textsc{ROLL}~\citep{wang2025reinforcement} & \textcolor{red}{$\times$} & \textcolor{red}{$\times$} & \textcolor{red}{$\times$} & \textcolor{red}{$\times$} & \textcolor{red}{$\times$} & \textcolor{red}{$\times$} \\
\textsc{RAGEN}~\citep{Wang2025RAGENUS} & \textcolor{red}{$\times$} & \textcolor{red}{$\times$} & \textcolor{red}{$\times$} & \textcolor{red}{$\times$} & \textcolor{red}{$\times$} & \textcolor{red}{$\times$} \\
\textsc{Slime}~\citep{slime2024} & \textcolor{searchgreen}{\checkmark} & \textcolor{red}{$\times$} & \textcolor{red}{$\times$} & \textcolor{red}{$\times$} & \textcolor{red}{$\times$} & \textcolor{red}{$\times$} \\
\textsc{AReal}~\citep{fu2025areal} & \textcolor{searchgreen}{\checkmark} & \textcolor{red}{$\times$} & \textcolor{searchgreen}{\checkmark} & \textcolor{red}{$\times$} & \textcolor{red}{$\times$} & \textcolor{red}{$\times$} \\
\textsc{SkyRL}~\citep{cao2025skyrl} & \textcolor{red}{$\times$} & \textcolor{searchgreen}{\checkmark} & \textcolor{red}{$\times$} & \textcolor{searchgreen}{\checkmark} & \textcolor{searchgreen}{\checkmark} & \textcolor{red}{$\times$} \\
\ourscolored (ours) & \textcolor{searchgreen}{\checkmark} & \textcolor{searchgreen}{\checkmark} & \textcolor{searchgreen}{\checkmark} & \textcolor{searchgreen}{\checkmark} & \textcolor{searchgreen}{\checkmark} & \textcolor{searchgreen}{\checkmark} \\
\bottomrule
\end{tabular}
\end{table}

The success of Reinforcement Learning from Verifier Rewards (RLVR) has motivated the development of various frameworks to support scalable RL training for language models. Established synchronous frameworks include \textsc{OpenRLHF}~\citep{hu2024openrlhf} and \textsc{VeRL}~\citep{sheng2024hybridflow}, which employ Ray-based distributed computing to manage training workflows. Additionally, fully asynchronous frameworks such as \textsc{AReal}~\citep{fu2025areal}, \textsc{ROLL}~\citep{wang2025reinforcement}, and \textsc{Slime}~\citep{slime2024} have emerged to address scalability challenges.

As shown in Table~\ref{tab:tool_supports}, existing frameworks exhibit varying degrees of tool support. Traditional RL frameworks such as \textsc{OpenRLHF} and \textsc{VeRL} provide basic support for search and code execution tools but lack comprehensive multi-modal capabilities. \textsc{ROLL} focuses primarily on core RL training without extensive tool integration, while \textsc{AReal} supports search functionality but has limited executor capabilities. \textsc{SkyRL}~\citep{cao2025skyrl} offers broader tool support, including bash terminals and SQL executors, but requires complex containerized environments that introduce deployment overhead. The limited tool coverage in existing frameworks has led to the development of domain-specific systems (e.g., \textsc{Search-R1}, \textsc{PixelReasoner}, and \textsc{ToolRL}) as task-specific extensions. However, these implementations typically feature hard-coded tool integrations that limit their extensibility and adaptability to new domains. As evident from Table~\ref{tab:tool_supports}, there remains a need for frameworks that provide comprehensive, extensible support for diverse tool types while maintaining ease of deployment and development.

\section{VerlTool Framework}
\label{sec:verltool}

In this section, we formulate the conceptual foundation of the ARLT paradigm starting from the original RLVR setting. We then elaborate on how \ours{} serves as a practical implementation on the server side for \textbf{A}gentic \textbf{R}einforcement \textbf{L}earning with \textbf{T}ool use (\textbf{ARLT}).

\subsection{Preliminaries}
\label{subsec:preliminary}

\paragraph{RLVR} Reinforcement learning with verifiable reward (RLVR) optimizes the language model using a predefined verifiable reward via the following objective:
\begin{equation}
\label{eq:rl_obj}
\max_{\pi_\theta} \mathbb{E}_{x \sim \mathcal{D}, y \sim \pi_\theta(\cdot \mid x)} 
\left[ R_\phi(x, y) \right] 
- \beta \, \mathbb{D}_{\text{KL}} \left[ \pi_\theta(y \mid x) \,\|\, \pi_{\text{ref}}(y \mid x) \right],
\end{equation}
where $\pi_{\theta}$ denotes the policy LLM, $\pi_{\text{ref}}$ is the reference LLM, $R_{\phi}$ is the verifiable reward function, and $\mathbb{D}_{\text{KL}}$ is the KL divergence. The input $x$ is drawn from the dataset $\mathcal{D}$, and $y$ is the corresponding single-turn output. A typical verifiable reward function is defined as:
\begin{equation}
    R_\phi(x,y) = \begin{cases}
1 & \text{if } \text{match}(y, y_g) \\
-1 & \text{otherwise}
\end{cases}
\end{equation}
where $y_g$ is the ground-truth answer and $\text{match}(\cdot, \cdot) \in \{1, 0\}$ is a verification function that determines whether the generated answer $y$ matches $y_g$. This function can be implemented using either rule-based approaches~\citep{Wang2024OpenHandsAO} or model-based verifiers~\citep{Ma2025GeneralReasonerAL}. 

\textbf{GRPO}~\citep{Shao2024DeepSeekMathPT} is a widely adopted RL algorithm designed to optimize the objective in~\autoref{eq:rl_obj}. In the single-turn RL case, the trajectory is simply the LLM generation $\tau = \{y\}$. The GRPO objective is given by:
\begin{equation}
    J_{\text{GRPO}}(\theta) = \frac{1}{G} \sum_{i=1}^G \frac{1}{|\tau_i|} \sum_{t=1}^{|\tau_i|} 
    \min \left[
        r_{i,t}(\theta) \cdot \hat{A}_{i,t},\,
        \text{clip}\left( r_{i,t}(\theta), 1 - \epsilon, 1 + \epsilon \right) \cdot \hat{A}_{i,t}
    \right],
\end{equation}
where $r_{i,t}(\theta)$ is the token-level importance ratio and $\hat{A}_{i,t}$ is the normalized advantage across all tokens:
\begin{equation}
    r_{i,t}(\theta) = \frac{\pi_\theta(\tau_{i,(t)}|\tau_{i,<t})}{\pi_{\text{old}}(\tau_{i,(t)}|\tau_{i,<t})}, \quad
    \hat{A}_{i,t} = \frac{R_\phi(x, y) - \text{mean}(\{R_\phi(\tau_1), \ldots, R_\phi(\tau_G)\})}
                         {\text{std}(\{R_\phi(\tau_1), \ldots, R_\phi(\tau_G)\})}.
\end{equation}

\subsection{Agentic Reinforcement Learning with Tool Use}
\label{subsec:agentrl}

\paragraph{ARLT.} In the agentic RL setting, rollouts are \textit{multi-turn} instead of single-turn, and the agent can \textit{interact with tools} to receive external observations during the reasoning process. Thus, the trajectory can be written as $\tau = \{a_0, o_0, \ldots, a_{n-1}, o_{n-1}, a_n\}$, where $a_i$ denotes the LLM-generated action tokens and $o_i$ denotes the observation tokens returned by a tool call. Here, $n$ is the total number of interaction steps.

To determine whether an action $a_i$ invokes a specific tool, we assume that each $a_i$ (for $0 \leq i < n$) ends with a stop token $s \in \mathbb{S}_k$, where $\mathbb{S}_k$ is the predefined set of stop tokens for tool $T_k \in \mathbb{T}$. For example, $\mathbb{S}_{\text{CI}} = \{\texttt{\texttt{`}\texttt{`}\texttt{`}output}, \texttt{</python>}\}$ for a code interpreter tool, or $\mathbb{S}_{\text{search}} = \{\texttt{</search>}\}$ for a search tool. The complete set of stop tokens is the union over all invoked tools: $\mathbb{S} = \bigcup_{k=1}^{|\mathbb{T}|} \mathbb{S}_k$.

The introduction of observation tokens $o_i$ makes ARLT fundamentally different from the agentic RL defined in \textsc{RAGEN}~\citep{Wang2025RAGENUS}, where the agent only receives scalar rewards through environmental interaction. Moreover, the observation tokens are off-policy with respect to the current LLM $\pi_\theta$ being optimized, which can destabilize training~\citep{searchr1}. Therefore, these tokens are typically masked out during policy optimization. Let $T_j$ be the token index of the first token in action segment $a_j$, then the GRPO loss for ARLT becomes:
\begin{equation}
    J_{\text{GRPO-ARLT}}(\theta) = \frac{1}{G} \sum_{i=1}^G \frac{1}{\sum_{j=0}^{n} |a_j|} 
    \sum_{j=0}^{n}
    \sum_{t=T_j}^{T_j + |a_j|} 
    \min \left[
        r_{i,t}(\theta) \cdot \hat{A}_{i,t},\,
        \text{clip}\left( r_{i,t}(\theta), 1 - \epsilon, 1 + \epsilon \right) \cdot \hat{A}_{i,t}
    \right],
\end{equation}


\subsection{Framework Design}
\label{subsec:framework}




\paragraph{Challenges.}
Building a general RL training framework that supports various tools is inherently challenging due to the additional overhead introduced by tool interactions. 
First, prior ARLT works are designed around a single tool and tightly couple the tool interaction logic with the core RL training loop, making it difficult for developers to extend or substitute tools~\citep{Li2025ToRLST}. This fragmentation increases the development burden for researchers seeking to experiment with novel tools or multi-tool scenarios. Second, these systems often rely on synchronous rollout mechanisms that process trajectories batch by batch~\citep{searchr1}. In such settings, the tool interaction phase is triggered only after all actions $a_i$ in a batch have been generated, resulting in idle bubbles and inefficient utilization of computational resources, especially on GPUs.

To address these issues, we propose \ours{}, a general-purpose ARLT framework designed to support various tools as modular plugins via a unified API. Our goal is to minimize the integration overhead for community developers and provide a more efficient and extensible infrastructure for training LLMs with tool-use capabilities.







\paragraph{Overview.} 
As shown in~\autoref{fig:overview}, \ourscolored{}  adopts a modular and decoupled architecture consisting of two main components: the {\color{cc2}\textbf{VeRL Workflow}} and the {\color{cc1}\textbf{Tool Server}}, connected via a unified API. This separation enables independent management of RL training and tool execution while preserving efficient communication between them.

The {\color{cc2}\textbf{VeRL Workflow}} handles all reinforcement learning activities, including multi-turn rollouts and actor updates. The LLM actor interacts with the external environment by generating a sequence of actions $\{a_0, a_1, \dots\}$, each potentially triggering a tool interaction. Once an action is identified as tool-invoking (via matching a predefined stop token), it is sent to the {\color{cc1}\textbf{Tool Server}} along with auxiliary metadata. The observation $o_i$ returned by the tool is then appended to the rollout, enabling observation-aware agent behavior and reward computation.



\begin{figure}[!t]
    \centering
    \begin{minipage}[!h]{0.48\textwidth}
        \includegraphics[width=\linewidth]{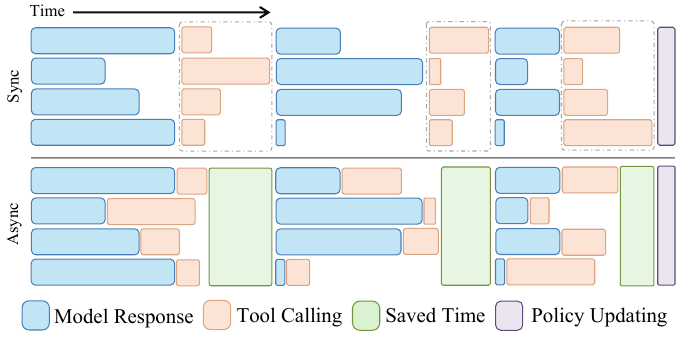}
        \caption{Visualization of the Async Rollout pipeline design and its effect in saving time.}
        \label{fig:asyncvis}
    \end{minipage}
    \hfill
    \begin{minipage}[!h]{0.48\textwidth}
        \vspace{0pt}
        \small
        \captionof{table}{Performance comparison of Synchronous vs Asynchronous approaches. Experiments conducted on 8 H100 GPUs.}
        \phantomcaption\label{tab:async_speedup}
        \begin{tabular}{lccc}
        \toprule
       Metrics & Math-TIR & SQL & DeepSearch \\
        \midrule
        Turns & 4 & 5 & 5 \\
        Sync (s) & 87 & 111 & 193 \\
        Async (s) & 66 & 91 & 98 \\
        \textbf{Speed Up ($\times$)} & \textbf{1.32} & \textbf{1.22} & \textbf{1.97} \\
        \bottomrule
        \end{tabular}
    \end{minipage}
\end{figure}

\paragraph{Asynchronous Rollout Design.}
A key feature of \ours{} is its support for fully \textit{asynchronous rollouts}, which avoids the inefficiency of traditional synchronous batch-based frameworks. In such a setting, tool calls are processed only after the entire batch has completed generating their respective actions $a_i$, resulting in idle "bubbles" in GPU and CPU utilization. In contrast, \ours{} enables each trajectory to interact with the tool server independently and immediately upon finishing its action generation, as shown in ~\autoref{fig:asyncvis}. This design ensures that tool execution latency does not block the entire batch, significantly improving throughput and system utilization in large-scale distributed settings. As shown in ~\hyperref[tab:async_speedup]{Table~\ref*{tab:async_speedup}}, the actor and environment evolve concurrently, achieving near $2$ times speedup for the rollout stage.


\paragraph{Modular Tool-as-Plugin Design.}

\begin{figure}[!t]
    \centering
    \includegraphics[width=0.9\textwidth]{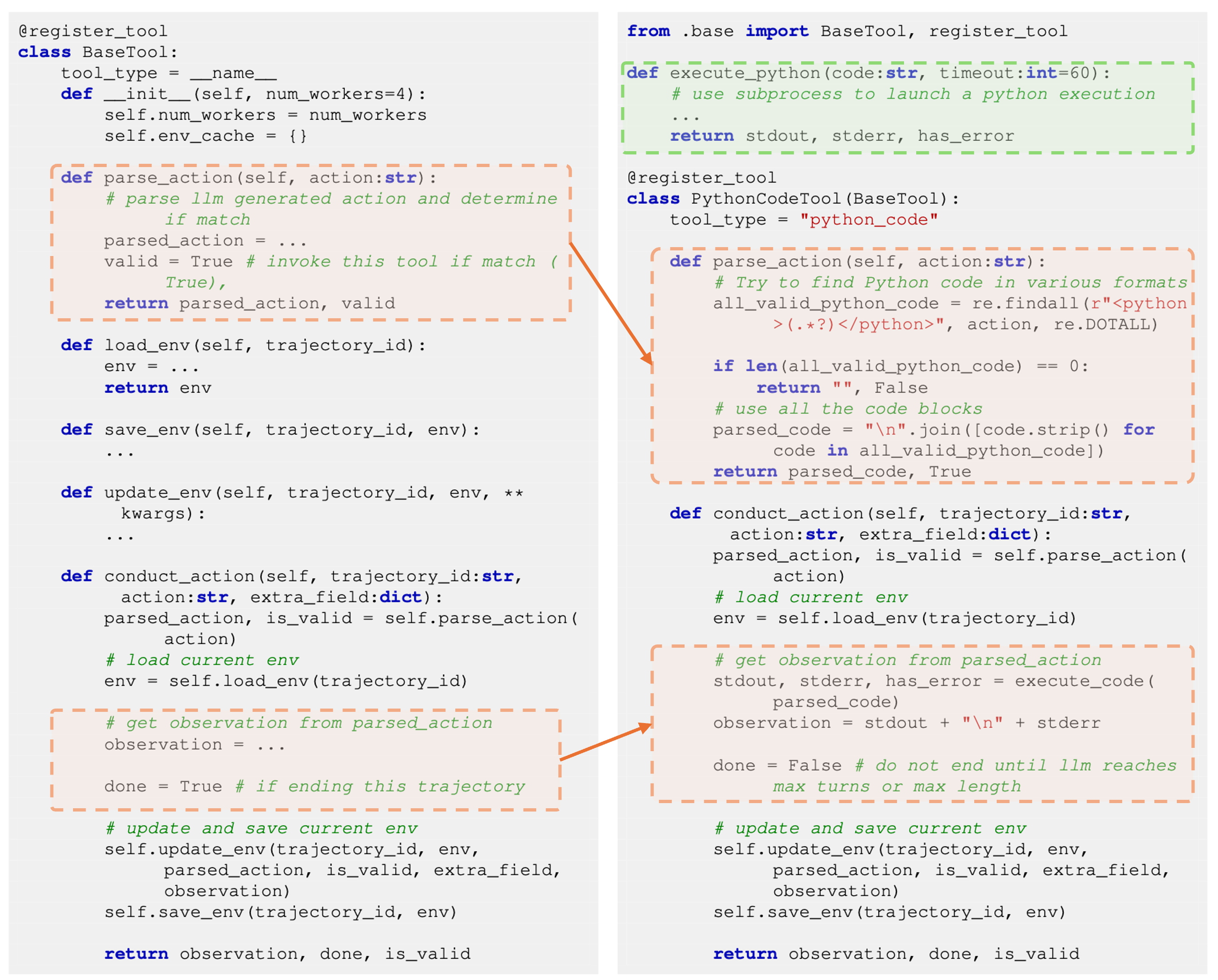}
    \caption{Example of code design for adding a new tool in \ours{} via the plugin interface.}
    \label{fig:adding_tool_example}
\end{figure}

As illustrated in ~\autoref{fig:adding_tool_example}, \ours{} adopts a modular plugin system that cleanly abstracts tool interaction as an interface between the LLM actor and its external environment. Each tool is implemented as a subclass of a unified \texttt{BaseTool}, enabling seamless registration and extensibility. During rollouts, the actor’s action $a_i$ is parsed by \texttt{parse\_action} to determine whether it invokes a tool; valid calls are routed to the appropriate module, which retrieves the trajectory state via \texttt{load\_env}. The tool then executes its \texttt{conduct\_action}, returning the observation $o_i$, a validity flag, and a termination flag for next action generation. We also maintain per-trajectory environments through lightweight state dictionaries, updated via \texttt{update\_env} and cleared at the end of an episode with \texttt{delete\_env}. By decoupling tool logic from the training workflow, developers can add new tools with minimal overhead, while the framework dynamically manages their execution across threads or distributed workers. 

\begin{wrapfigure}{r}{0.43\textwidth}
    \centering
    \vspace{-2em}
    \includegraphics[width=0.40\textwidth]{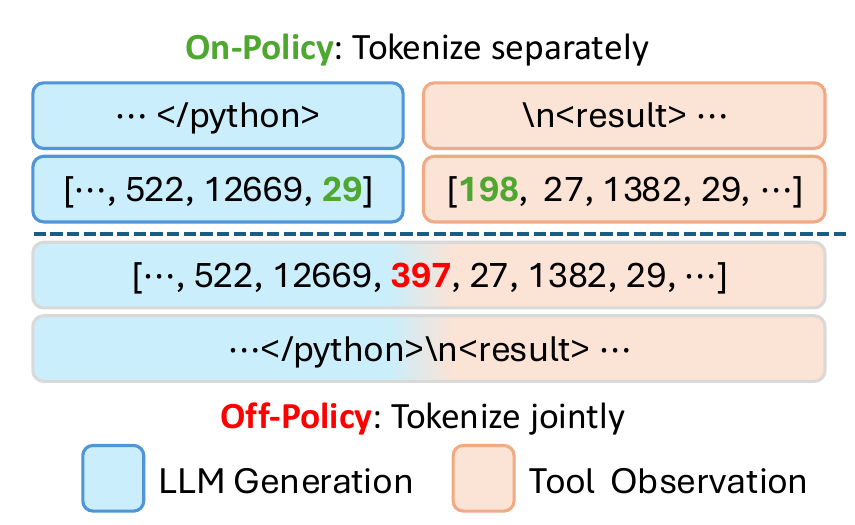}
    \caption{Tokenization of LLM generated content ``...\texttt{</python>}'' and tool observation ``\texttt{\textbackslash n<result>}...'' can produce different token lists using Qwen2.5 tokenizer under different strategies.}
    \label{fig:tokenization}
    \vspace{-30pt}
\end{wrapfigure}

\paragraph{Tokenization.} 
A practical challenge in multi-turn agentic RL is how to tokenize tool observations and concatenate them with preceding LLM actions. Two strategies exist: (i) tokenize the action and observation strings separately, then append their token sequences; or (ii) concatenate the raw strings first and tokenize jointly. Although they yield the same sequence most of the time, discrepancies may arise for some specific combinations, as illustrated in \autoref{fig:tokenization}. 
The action string ``\texttt{</python>}'' and observation ``\texttt{\textbackslash n<result>}'' produce consistent tokens under the first strategy (\emph{On-Policy}), whereas the second one merges boundary symbols into a different token id (e.g., \texttt{29,198} vs.\ \texttt{397}), which changed the LLM-generated contents (\emph{Off-Policy}). 
To avoid such inconsistencies, we adopt the first approach and always keep a consistent token list prefix during rollout, 
ensuring stable alignment in multiple rollouts turns.

\paragraph{Parallel Tool Server Backend.}
To support high-throughput and scalable execution of tool interactions, \ours{} offers two types of parallel execution backends within the Tool Server:
\begin{itemize}
\setlength{\itemsep}{0pt}
\setlength{\parskip}{0pt}
    \item \emph{Multi-threading}: For small-scale or lightweight tool calls, \ours{} employs Python's \texttt{ThreadPoolExecutor} to parallelize calls across multiple worker threads.
    \item \emph{Ray-based Asynchronous Execution}: To deal with resource-intensive tools for better resource management, \ours{} optionally supports integration with Ray~\citep{Moritz2017RayAD}, enabling distributed and fault-tolerant tool execution across machines or GPU nodes. This design provides robust scalability for long-horizon or computationally intensive tools.
\end{itemize}

\section{Experiments}
\label{sec:exp}
\begin{figure}[!t]
    \centering
    \includegraphics[width=\linewidth]{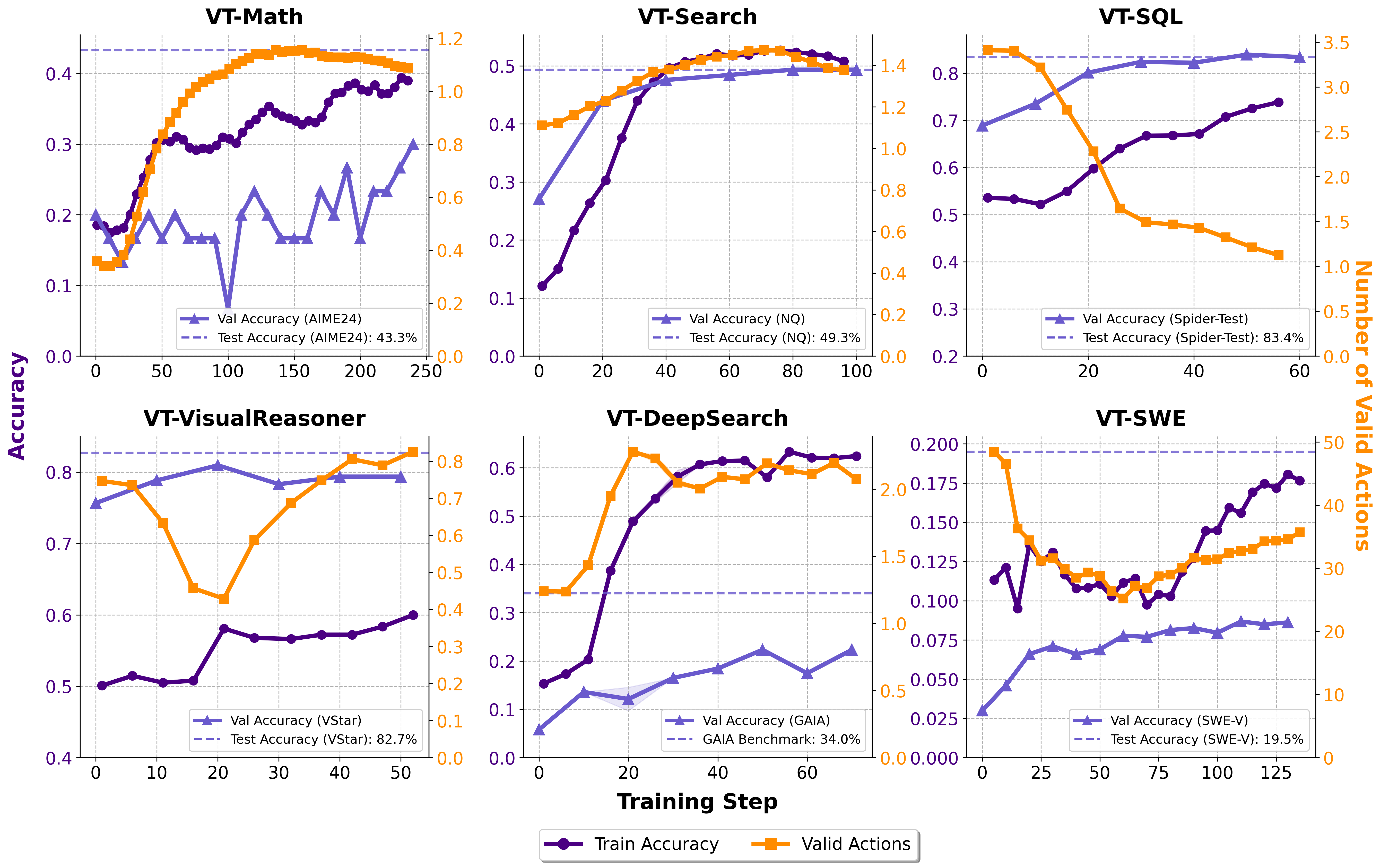}
    \caption{Training dynamics using \ours{} on all 6 tasks. For each task, the corresponding test benchmarks are AIME24, NQ, Spider-Test, VStar, GAIA, and SWE-Verified. All models are trained and evaluated based on \ourscolored{} framework. The actual evaluation performance (purple dash) can be higher due to the train-eval settings difference. The number of actions is averaged over all sampled responses in each batch.}
    \label{fig:train_dynamics}
    \vspace{-1em}
\end{figure}

\subsection{Experiment Setup}
With a modular, plug-and-play design, \ours{} equips an agent with tools spanning multiple domains and modalities as shown in ~\autoref{tab:all_tools}. In this section, we show the experiment results in six agentic RL with tool use (ARLT) tasks, including VT-Math (\autoref{tab:res_math}), VT-Search (\autoref{tab:res_searchr1}), VT-SQL (\autoref{tab:sql_swe}), VT-VisualReasoner (\autoref{tab:res_ds_pr}), VT-DeepSearch (\autoref{tab:res_ds_pr}), and VT-SWE (\autoref{tab:sql_swe}), demonstrating the compatibility of \ours{} with various tools. Please see details of training, evaluation in ~\autoref{sec:exp_setup}.

\subsection{Results}

\begin{table}[!t]
    \centering
    \renewcommand{\arraystretch}{1.2}
    \setlength{\tabcolsep}{2pt}
    \vspace{-0.3em}
    \caption{Results on Math-related benchmarks with Python interpreter tool.
    The best results are indicated in \textbf{bold} and the second-best results are \underline{underlined}. 
    }
    \label{tab:res_math}
    \vspace{-1em}
    \resizebox{\columnwidth}{!}{
    \begin{tabular}{lccccccc}
    \toprule
    \textbf{Model} & \textbf{GSM8K} & \textbf{MATH 500} & \textbf{Minerva Math} & \textbf{Olympiad Bench} & \textbf{AIME24} & \textbf{AMC23} & \textbf{Avg.} \\
    \midrule

    
    \multicolumn{8}{c}{\textit{\textbf{Qwen2.5-Math-7B-Base/Instruct}}} \\
    \midrule
    Qwen2.5-Math-7B-Instruct        & \textbf{95.2} & \underline{83.0} & \textbf{37.1} & 41.6 & 16.7 & 70.0 & 57.3 \\
    Qwen-2.5-Math-7B-Instruct-TIR   & 88.8 & 80.2 & 26.8 & 41.6 & 30.0 & 52.5 & 53.3 \\
    SimpleRL-Zoo-7B                 & \underline{94.6} & 82.4 & 29.0 & \underline{50.5} & 30.0 & 62.5 & 58.2 \\
    ToRL-7B                         & 92.7 & 82.2 & 33.5 & 49.9 & \underline{43.3} & 65.0 & 61.1 \\
    \hdashline
    
    VT-Math-zero (GRPO)             & 91.8 & \textbf{83.2} & 31.6 & \underline{50.5} & \textbf{43.3} & \underline{70.0} & \underline{61.7} \\
    VT-Math-zero (DAPO)             & 92.1 & 82.8 & \underline{34.9} & \textbf{51.6} & 36.7 & \textbf{75.0} & \textbf{62.2} \\
    
    \bottomrule
    \end{tabular}}
\end{table}

\begin{table}[!t]
    \centering
    \setlength{\tabcolsep}{5pt}
    \renewcommand{\arraystretch}{1.2}
    \vspace{-0.2em}
    
    \caption{Results of on knowledge-QA benchmarks. $^\dagger/^\star$ represents in-domain/out-domain datasets. 
    }
    \label{tab:res_searchr1}
    \vspace{-1em}
    \resizebox{\columnwidth}{!}{
    \begin{tabular}{lcccccccc}
        \toprule
        \textbf{Model} & \multicolumn{3}{c}{\textbf{General QA}} & \multicolumn{4}{c}{\textbf{Multi-Hop QA}} & \textbf{Avg.} \\
        \cmidrule(lr){2-4}\cmidrule(lr){5-8}\cmidrule(lr){9-9}
         & \textbf{NQ$^\dagger$} & \textbf{TriviaQA$^\star$} & \textbf{PopQA$^\star$}
         & \textbf{HotpotQA$^\dagger$} & \textbf{2wiki$^\star$} & \textbf{Musique$^\star$} & \textbf{Bamboogle$^\star$}
         & \textbf{Avg.} \\


        \midrule
        
        \multicolumn{9}{c}{\textbf{\textit{Qwen2.5-7b-Base/Instruct}}} \\
        \midrule
        Direct Inference         & 13.4 & 40.8 & 14.0 & 18.3 & 25.0 & 3.1 & 12.0 & 18.1 \\
        Search-R1-base (GRPO)    & 39.5 & 56.0 & 38.8 & 32.6 & 29.7 & 12.5 & 36.0 & 35.0 \\
        Search-R1-base (PPO)     & 48.0 & \underline{63.8} & 45.7 & \underline{43.3} & 38.2 & \textbf{19.6} & \underline{43.2} & \underline{43.1} \\
        \hdashline
        VT-Search-zero (GRPO)    & \textbf{49.3} & \textbf{66.2} & \textbf{50.2} & \textbf{44.8} & \textbf{45.3} & \underline{19.3} & \textbf{46.4} & \textbf{45.9} \\
        VT-Search-zero (DAPO)    & \underline{48.3} & 63.4 & \underline{48.2} & 42.6 & \underline{39.2} & 18.0 & 38.4 & 41.2 \\
        \bottomrule
    \end{tabular}
    }
\end{table}

\begin{table}[!t]
\centering
\vspace{-0.2em}
\caption{
    Results on NL2SQL (left) and SWE-Verified (right) benchmarks in terms of pass rates.
}
\label{tab:sql_swe} 
\vspace{-1em}
\begin{minipage}{0.48\textwidth}
\centering
\small
\setlength{\tabcolsep}{6pt}
\renewcommand{\arraystretch}{1.2}
\selectfont
\resizebox{\textwidth}{!}{
\begin{tabular}{lcc}
   \toprule
   \textbf{Model} & \multicolumn{2}{c}{\textbf{Spider}} \\
   \textbf{Split} & \textbf{Dev} & \textbf{Realistic} \\
    \midrule
    \multicolumn{3}{c}{\textbf{\textit{Reasoning without Tool}}} \\
    \midrule
    GPT-4o                      & 70.9 & - \\
    DeepSeekCoder-6.7B-Instruct           & 63.2 & - \\
    OpenCoder-8B-Instruct       & 59.5 & - \\
    Qwen2.5-Coder-7B-Instruct   & 73.4 & - \\
    \midrule
    \multicolumn{3}{c}{\textbf{\textit{Tool Integrated Reasoning}}} \\
    \midrule
    OmniSQL-7B                  & \underline{81.2} & 63.9 \\
    SkyRL-SQL-7B (GRPO)         & \textbf{83.9} & \underline{81.1} \\
    \hdashline
    \multicolumn{3}{l}{{\textit{VT-SQL (Qwen-2.5-Coder-7B-Instruct based)}}} \\
    \quad + GRPO                & \textbf{83.9} & \textbf{81.3} \\
\bottomrule
\end{tabular}}
\end{minipage}%
\hfill
\begin{minipage}{0.45\textwidth}
\centering
\small
\setlength{\tabcolsep}{6pt}
\renewcommand{\arraystretch}{1.2}
\selectfont
\resizebox{\textwidth}{!}{
\begin{tabular}{lc}
\toprule
\textbf{Model} & \textbf{SWEBench} \\
\addlinespace[0.2em]
\midrule
\addlinespace[0.3em]
\multicolumn{2}{c}{\textit{\textbf{OpenHands Scaffold}}} \\
\addlinespace[0.2em]
\midrule
Qwen3-8B             & 3.6 \\
OpenHands-7B-Agent             & 11.0 \\
\addlinespace[0.3em]
\hdashline
\addlinespace[0.3em]
\textit{SkyRL-v0} \citep{cao2025skyrl} \\
\quad Qwen3-8B Based & 9.4 \\
\quad OpenHands-7B-Agent Based & 14.6 \\
\midrule
\multicolumn{2}{c}{\textit{\textbf{R2E Gym Scaffold}}} \\
\midrule
Qwen3-8B             & 10.4 \\
\addlinespace[0.3em]
\hdashline
\addlinespace[0.3em]
\textit{VT-SWE (Qwen3-8B Based)} \\
\quad +GRPO & \textbf{19.5} \\
\bottomrule
\end{tabular}
}
\end{minipage}
\end{table}
\begin{table}[!h]
\centering
\small
\vspace{-0.25em}
\caption{ Results on visual reasoning (left) and agentic search benchmarks (right).}
\label{tab:res_ds_pr}
\vspace{-1em}
\renewcommand{\arraystretch}{1.2}
\setlength{\tabcolsep}{4pt}

\begin{minipage}{0.48\textwidth}
\centering
\begin{tabular}{lc}
\toprule
\textbf{Model} & \textbf{V* Bench} \\
\midrule
\multicolumn{2}{c}{\textit{\textbf{Reasoning without Tool}}} \\
\midrule
GPT-4o & 62.8 \\
Gemini-2.5-Pro & 79.2 \\
Qwen2.5-VL-7B-Instruct & 70.4 \\
Video-R1-7B & 51.2 \\
\midrule
\multicolumn{2}{c}{\textit{\textbf{Tool Integrated Reasoning}}} \\
\midrule
Visual Sketchpad (GPT-4o) & 80.4 \\
IVM-Enhanced (GPT-4V) & 81.2 \\
Pixel-Reasoner-7B & \textbf{84.3} \\
\hdashline
\multicolumn{2}{l}{\textit{VT-VisualReasoner (Qwen2.5-VL-7B-Instruct Based)}} \\
\quad + GRPO-Acc & 78.8 \\
\quad + GRPO-Complex & \underline{82.7} \\
\bottomrule
\end{tabular}
\end{minipage}
\hfill
\begin{minipage}{0.48\textwidth}
\centering
\begin{tabular}{lcc}
\toprule
\textbf{Model} & \textbf{GAIA} & \textbf{HLE} \\
\midrule
\multicolumn{3}{c}{\textit{\textbf{Reasoning without Tool}}} \\
\midrule
DeepSeek-R1-671B   & \underline{25.2} & \textbf{8.6} \\
GPT-4o             & 17.5 & 2.6 \\
Qwen3-8B           & 20.4 & 4.6 \\
\midrule
\multicolumn{3}{c}{\textit{\textbf{Tool Integrated Reasoning (Qwen3-8B)}}} \\
\midrule
Vanilla RAG        & 20.4 & 5.8 \\
Search-o1          & 21.4 & 6.4 \\
WebThinker         & 22.3 & 6.6 \\
ReAct              & 23.3 & 4.6 \\
\hdashline
\multicolumn{3}{l}{\textit{VT-DeepSearch (Qwen3-8B Based)}} \\
\quad + Snippets-Only    & 32.0 & 7.8 \\
\quad + QwQ-32B    & \textbf{34.0} & \underline{8.4} \\
\bottomrule
\end{tabular}
\end{minipage}

\vspace{-1em}
\end{table}

\paragraph{Training on \ours{} achieves competitive results.} Models trained using \ours{} consistently match or exceed existing baselines across all six tasks. VT-Math achieves a 62.2\% average performance on mathematical benchmarks, surpassing expert models on multiple benchmarks such as AIME24, AMC23, and Olympiad Bench. VT-Search reaches 45.9\% accuracy on knowledge QA, surpassing Search-R1 by 10.9\%. In terms of the NL2SQL task, VT-SQL matches specialized systems such as SkyRL-SQL. VT-VisualReasoner achieves 82.7\% on V* Bench while VT-DeepSearch reaches 34.0\% on GAIA. These figures demonstrate that trained under our unified framework, agents could achieve competitive task-specific performance compared to separate divergent code bases.

\paragraph{Multi-modal tools are well-supported.} \ours{}'s modular design enables the seamless integration of a wide range of types of tools within a unified API interface. We demonstrated this capability through the implementation of text-based tools support including Python and SQL Interpreter, Local Retriever and Web Search Tools, as well as visual processing tools that operate on image and video modality (image operations and video frame selections). Moreover, our framework is equipped with system-level tools such as the bash terminal and file operation tools. Visual reasoning experiments demonstrate that our agents could dynamically manipulate images and process visual information iteratively, enabling complex multi-modal workflows that were unsupported in existing single-modality RL-training frameworks, which only work on the text level.

\paragraph{Dynamics of Tool Usage across Tasks.} 
Tool usage patterns exhibit substantial variation across different domains, with mathematical tasks typically requiring $1 \sim 4$ interactions while software engineering tasks may extend to over 100 interactions. Importantly, models do not spontaneously develop effective tool-use capabilities without appropriate reward design and initialization strategies. For instance, without VT-VisualReasoner's sophisticated reward mechanism, the frequency of tool actions gradually diminishes to zero within a few reinforcement learning (RL) steps.

The evolution of tool usage during training demonstrates task-specific characteristics that reflect the underlying utility of tool interactions. In VT-SQL settings, the number of actions decreases rapidly after several dozen RL steps, as the model learns that SQL executors are non-essential for most straightforward queries. Through training, the model develops a preference for responses requiring fewer tool calls by gradually memorizing expected execution results, thereby reducing the need for verification through SQL executors. Conversely, in VT-DeepSearch settings, tool usage increases dramatically during training because problem-solving fundamentally depends on search capabilities. Unlike simpler SQL results, the extensive and information-rich content returned by search tools cannot be easily memorized, necessitating increased tool invocations for effective task completion. This divergent behavior underscores how the intrinsic value of tool assistance shapes learning dynamics across different computational domains.

\paragraph{Emerging behaviours of Agentic RL.} Agentic models trained using \ours{} emerged sophisticated behaviors, including self-correction, iterative refinement, and strategic tool selection in the multi-round tool-calling environment provided by our framework. For instance, mathematical agents verify computations and backtrack from errors, search agents refine queries based on retrieved information, and software agents develop debugging strategies combining code analysis and incremental fixes. These capabilities represent genuine agentic problem-solving that extends beyond simple function-calling invocations. We present corresponding case studies in ~\autoref{sec:case_study}.

\section{Conclusion}

We propose \ours{}, addressing key limitations of Agentic Reinforcement Learning with Tool use (ARLT) models' training. Our framework features a unified and modular systematic design, providing multi-modal tool management through standardized API designs, while maintaining high-efficiency model training featuring asynchronous rollout execution. Our system extends traditional single-turn reinforcement learning with verifiable rewards to ARLT domains, featuring robust system designs and upstream-aligned with VeRL. The framework is extensively examined across six domains featuring diverse tool integrations and modalities. As evidenced by extensive evaluation, agents trained through our framework demonstrated competitive performance compared to specialized systems, while unified under our training infrastructure. We present \ours{} as a scalable foundational training infrastructure to the RL community and hope our contributions could facilitate the advancement of ARLT research.


\bibliography{iclr2026_conference}

\begin{thebibliography}{104}
\providecommand{\natexlab}[1]{#1}
\providecommand{\url}[1]{\texttt{#1}}
\expandafter\ifx\csname urlstyle\endcsname\relax
  \providecommand{\doi}[1]{doi: #1}\else
  \providecommand{\doi}{doi: \begingroup \urlstyle{rm}\Url}\fi

\bibitem[Cai et~al.(2021)Cai, Yuan, Xu, and Hao]{Cai2021SADGASD}
Ruichu Cai, Jinjie Yuan, Boyan Xu, and Zhifeng Hao.
\newblock Sadga: Structure-aware dual graph aggregation network for text-to-sql.
\newblock \emph{ArXiv}, abs/2111.00653, 2021.
\newblock URL \url{https://api.semanticscholar.org/CorpusID:240353884}.

\bibitem[Cao et~al.(2025)Cao, Hegde, Li, Griggs, Liu, Tang, Pan, Wang, Malik, Neubig, Hakhamaneshi, Liaw, Moritz, Zaharia, Gonzalez, and Stoica]{cao2025skyrl}
Shiyi Cao, Sumanth Hegde, Dacheng Li, Tyler Griggs, Shu Liu, Eric Tang, Jiayi Pan, Xingyao Wang, Akshay Malik, Graham Neubig, Kourosh Hakhamaneshi, Richard Liaw, Philipp Moritz, Matei Zaharia, Joseph~E. Gonzalez, and Ion Stoica.
\newblock Skyrl-v0: Train real-world long-horizon agents via reinforcement learning, 2025.

\bibitem[Chen et~al.(2022)Chen, Ma, Wang, and Cohen]{Chen2022ProgramOT}
Wenhu Chen, Xueguang Ma, Xinyi Wang, and William~W. Cohen.
\newblock Program of thoughts prompting: Disentangling computation from reasoning for numerical reasoning tasks.
\newblock \emph{Trans. Mach. Learn. Res.}, 2023, 2022.
\newblock URL \url{https://api.semanticscholar.org/CorpusID:253801709}.

\bibitem[Chen et~al.(2024{\natexlab{a}})Chen, Xu, Liang, He, Pang, Yu, Song, Liu, Zhou, Zhang, Wang, Tu, Mi, and Yu]{Chen2024DoNT}
Xingyu Chen, Jiahao Xu, Tian Liang, Zhiwei He, Jianhui Pang, Dian Yu, Linfeng Song, Qiuzhi Liu, Mengfei Zhou, Zhuosheng Zhang, Rui Wang, Zhaopeng Tu, Haitao Mi, and Dong Yu.
\newblock Do not think that much for 2+3=? on the overthinking of o1-like llms.
\newblock \emph{ArXiv}, abs/2412.21187, 2024{\natexlab{a}}.
\newblock URL \url{https://api.semanticscholar.org/CorpusID:275133600}.

\bibitem[Chen et~al.(2025)Chen, Min, Zhang, Chen, Jiang, Cheng, Zhao, Liu, Miao, Lu, Fang, Wang, and Wen]{Slow_Thinking_with_LLMs_3_Tool}
Zhipeng Chen, Yingqian Min, Beichen Zhang, Jie Chen, Jinhao Jiang, Daixuan Cheng, Wayne~Xin Zhao, Zheng Liu, Xu~Miao, Yang Lu, Lei Fang, Zhongyuan Wang, and Ji-Rong Wen.
\newblock An empirical study on eliciting and improving r1-like reasoning models.
\newblock \emph{arXiv preprint arXiv:2503.04548}, 2025.

\bibitem[Chen et~al.(2024{\natexlab{b}})Chen, Ma, Zhang, Hao, Yan, Nourbakhsh, Yang, McAuley, Petzold, and Wang]{Chen2024ASO}
Zhiyu~Zoey Chen, Jing Ma, Xinlu Zhang, Nan Hao, An~Yan, Armineh Nourbakhsh, Xianjun Yang, Julian~J. McAuley, Linda~Ruth Petzold, and William~Yang Wang.
\newblock A survey on large language models for critical societal domains: Finance, healthcare, and law.
\newblock \emph{ArXiv}, abs/2405.01769, 2024{\natexlab{b}}.
\newblock URL \url{https://api.semanticscholar.org/CorpusID:269587715}.

\bibitem[Cobbe et~al.(2021)Cobbe, Kosaraju, Bavarian, Chen, Jun, Kaiser, Plappert, Tworek, Hilton, Nakano, Hesse, and Schulman]{cobbe2021gsm8k}
Karl Cobbe, Vineet Kosaraju, Mohammad Bavarian, Mark Chen, Heewoo Jun, Lukasz Kaiser, Matthias Plappert, Jerry Tworek, Jacob Hilton, Reiichiro Nakano, Christopher Hesse, and John Schulman.
\newblock Training verifiers to solve math word problems, 2021.
\newblock URL \url{https://arxiv.org/abs/2110.14168}.

\bibitem[Dong et~al.(2025)Dong, Mao, Ma, Bao, Chen, Wang, Chen, Du, Wang, Zhang, et~al.]{dong2025agentic}
Guanting Dong, Hangyu Mao, Kai Ma, Licheng Bao, Yifei Chen, Zhongyuan Wang, Zhongxia Chen, Jiazhen Du, Huiyang Wang, Fuzheng Zhang, et~al.
\newblock Agentic reinforced policy optimization.
\newblock \emph{arXiv preprint arXiv:2507.19849}, 2025.

\bibitem[Douze et~al.(2024)Douze, Guzhva, Deng, Johnson, Szilvasy, Mazar'e, Lomeli, Hosseini, and J'egou]{Douze2024TheFL}
Matthijs Douze, Alexandr Guzhva, Chengqi Deng, Jeff Johnson, Gergely Szilvasy, Pierre-Emmanuel Mazar'e, Maria Lomeli, Lucas Hosseini, and Herv'e J'egou.
\newblock The faiss library.
\newblock \emph{ArXiv}, abs/2401.08281, 2024.
\newblock URL \url{https://api.semanticscholar.org/CorpusID:267028372}.

\bibitem[Epstein \& Axtell(1996)Epstein and Axtell]{epstein1996growing}
Joshua~M Epstein and Robert Axtell.
\newblock \emph{Growing artificial societies: social science from the bottom up}.
\newblock Brookings Institution Press, 1996.

\bibitem[Feng et~al.(2025)Feng, Huang, Qu, Zhang, Qin, Zhong, Jiang, Chi, and Zhong]{Feng2025ReToolRL}
Jiazhan Feng, Shijue Huang, Xingwei Qu, Ge~Zhang, Yujia Qin, Baoquan Zhong, Chengquan Jiang, Jinxin Chi, and Wanjun Zhong.
\newblock Retool: Reinforcement learning for strategic tool use in llms, 2025.
\newblock URL \url{https://arxiv.org/abs/2504.11536}.

\bibitem[Fu et~al.(2025)Fu, Gao, Shen, Zhu, Mei, He, Xu, Wei, Mei, Wang, Yang, Yuan, and Wu]{fu2025areal}
Wei Fu, Jiaxuan Gao, Xujie Shen, Chen Zhu, Zhiyu Mei, Chuyi He, Shusheng Xu, Guo Wei, Jun Mei, Jiashu Wang, Tongkai Yang, Binhang Yuan, and Yi~Wu.
\newblock Areal: A large-scale asynchronous reinforcement learning system for language reasoning, 2025.
\newblock URL \url{https://arxiv.org/abs/2505.24298}.

\bibitem[Gan et~al.(2021{\natexlab{a}})Gan, Chen, Huang, Purver, Woodward, Xie, and Huang]{gan-etal-2021-towards}
Yujian Gan, Xinyun Chen, Qiuping Huang, Matthew Purver, John~Robert Woodward, Jinxia Xie, and Pengsheng Huang.
\newblock Towards robustness of text-to-sql models against synonym substitution.
\newblock \emph{ArXiv}, abs/2106.01065, 2021{\natexlab{a}}.
\newblock URL \url{https://api.semanticscholar.org/CorpusID:235293739}.

\bibitem[Gan et~al.(2021{\natexlab{b}})Gan, Chen, and Purver]{gan2021exploring}
Yujian Gan, Xinyun Chen, and Matthew Purver.
\newblock Exploring underexplored limitations of cross-domain text-to-sql generalization, 2021{\natexlab{b}}.

\bibitem[Gao et~al.(2023{\natexlab{a}})Gao, Wang, Li, Sun, Qian, Ding, and Zhou]{Gao2023TexttoSQLEB}
Dawei Gao, Haibin Wang, Yaliang Li, Xiuyu Sun, Yichen Qian, Bolin Ding, and Jingren Zhou.
\newblock Text-to-sql empowered by large language models: A benchmark evaluation.
\newblock \emph{Proc. VLDB Endow.}, 17:\penalty0 1132--1145, 2023{\natexlab{a}}.
\newblock URL \url{https://api.semanticscholar.org/CorpusID:261276437}.

\bibitem[Gao et~al.(2023{\natexlab{b}})Gao, Xiong, Gao, Jia, Pan, Bi, Dai, Sun, Guo, Wang, and Wang]{Gao2023RetrievalAugmentedGF}
Yunfan Gao, Yun Xiong, Xinyu Gao, Kangxiang Jia, Jinliu Pan, Yuxi Bi, Yi~Dai, Jiawei Sun, Qianyu Guo, Meng Wang, and Haofen Wang.
\newblock Retrieval-augmented generation for large language models: A survey.
\newblock \emph{ArXiv}, abs/2312.10997, 2023{\natexlab{b}}.
\newblock URL \url{https://api.semanticscholar.org/CorpusID:266359151}.

\bibitem[Gemini(2024)]{geminiteam2024geminifamilyhighlycapable}
Gemini.
\newblock Gemini: A family of highly capable multimodal models, 2024.
\newblock URL \url{https://arxiv.org/abs/2312.11805}.

\bibitem[Gilbert(2019)]{gilbert2019agent}
Nigel Gilbert.
\newblock \emph{Agent-based models}.
\newblock Sage Publications, 2019.

\bibitem[Gou et~al.(2023)Gou, Shao, Gong, Shen, Yang, Huang, Duan, and Chen]{gou2023tora}
Zhibin Gou, Zhihong Shao, Yeyun Gong, Yelong Shen, Yujiu Yang, Minlie Huang, Nan Duan, and Weizhu Chen.
\newblock Tora: A tool-integrated reasoning agent for mathematical problem solving.
\newblock \emph{arXiv preprint arXiv:2309.17452}, 2023.

\bibitem[Guo et~al.(2025)Guo, Yang, Zhang, Song, Zhang, Xu, Zhu, Ma, Wang, Bi, et~al.]{r1}
Daya Guo, Dejian Yang, Haowei Zhang, Junxiao Song, Ruoyu Zhang, Runxin Xu, Qihao Zhu, Shirong Ma, Peiyi Wang, Xiao Bi, et~al.
\newblock Deepseek-r1: Incentivizing reasoning capability in llms via reinforcement learning.
\newblock \emph{arXiv preprint arXiv:2501.12948}, 2025.

\bibitem[He et~al.(2024)He, Luo, Bai, Hu, Thai, Shen, Hu, Han, Huang, Zhang, et~al.]{he2024olympiadbench}
Chaoqun He, Renjie Luo, Yuzhuo Bai, Shengding Hu, Zhen~Leng Thai, Junhao Shen, Jinyi Hu, Xu~Han, Yujie Huang, Yuxiang Zhang, et~al.
\newblock Olympiadbench: A challenging benchmark for promoting agi with olympiad-level bilingual multimodal scientific problems.
\newblock \emph{arXiv preprint arXiv:2402.14008}, 2024.

\bibitem[He et~al.(2025)He, Liang, Xu, Liu, Chen, Wang, Song, Yu, Liang, Wang, Zhang, Wang, Tu, Mi, and Yu]{He2025DeepMath103KAL}
Zhiwei He, Tian Liang, Jiahao Xu, Qiuzhi Liu, Xingyu Chen, Yue Wang, Linfeng Song, Dian Yu, Zhenwen Liang, Wenxuan Wang, Zhuosheng Zhang, Rui Wang, Zhaopeng Tu, Haitao Mi, and Dong Yu.
\newblock Deepmath-103k: A large-scale, challenging, decontaminated, and verifiable mathematical dataset for advancing reasoning.
\newblock \emph{ArXiv}, abs/2504.11456, 2025.
\newblock URL \url{https://api.semanticscholar.org/CorpusID:277787455}.

\bibitem[Hendrycks et~al.(2021)Hendrycks, Burns, Kadavath, Arora, Basart, Tang, Song, and Steinhardt]{hendrycksmath2021}
Dan Hendrycks, Collin Burns, Saurav Kadavath, Akul Arora, Steven Basart, Eric Tang, Dawn Song, and Jacob Steinhardt.
\newblock Measuring mathematical problem solving with the {MATH} dataset.
\newblock In \emph{Thirty-fifth Conference on Neural Information Processing Systems Datasets and Benchmarks Track (Round 2)}, 2021.
\newblock URL \url{https://openreview.net/forum?id=7Bywt2mQsCe}.

\bibitem[Ho et~al.(2020)Ho, Nguyen, Sugawara, and Aizawa]{ho2020constructing}
Xanh Ho, Anh-Khoa~Duong Nguyen, Saku Sugawara, and Akiko Aizawa.
\newblock Constructing a multi-hop qa dataset for comprehensive evaluation of reasoning steps.
\newblock \emph{arXiv preprint arXiv:2011.01060}, 2020.

\bibitem[Hu et~al.(2024)Hu, Wu, Zhu, Xianyu, Wang, Zhang, and Cao]{hu2024openrlhf}
Jian Hu, Xibin Wu, Zilin Zhu, Xianyu, Weixun Wang, Dehao Zhang, and Yu~Cao.
\newblock Openrlhf: An easy-to-use, scalable and high-performance rlhf framework.
\newblock \emph{arXiv preprint arXiv:2405.11143}, 2024.

\bibitem[Huang et~al.(2023)Huang, Yu, Ma, Zhong, Feng, Wang, Chen, Peng, Feng, Qin, and Liu]{Huang2023ASO}
Lei Huang, Weijiang Yu, Weitao Ma, Weihong Zhong, Zhangyin Feng, Haotian Wang, Qianglong Chen, Weihua Peng, Xiaocheng Feng, Bing Qin, and Ting Liu.
\newblock A survey on hallucination in large language models: Principles, taxonomy, challenges, and open questions.
\newblock \emph{ACM Transactions on Information Systems}, 43:\penalty0 1 -- 55, 2023.
\newblock URL \url{https://api.semanticscholar.org/CorpusID:265067168}.

\bibitem[Huang et~al.(2022)Huang, Abbeel, Pathak, and Mordatch]{Huang2022LanguageMA}
Wenlong Huang, P.~Abbeel, Deepak Pathak, and Igor Mordatch.
\newblock Language models as zero-shot planners: Extracting actionable knowledge for embodied agents.
\newblock \emph{ArXiv}, abs/2201.07207, 2022.
\newblock URL \url{https://api.semanticscholar.org/CorpusID:246035276}.

\bibitem[Inaba et~al.(2023)Inaba, Kiyomaru, Cheng, and Kurohashi]{Inaba2023MultiToolCoTGC}
Tatsuro Inaba, Hirokazu Kiyomaru, Fei Cheng, and Sadao Kurohashi.
\newblock Multitool-cot: Gpt-3 can use multiple external tools with chain of thought prompting.
\newblock \emph{ArXiv}, abs/2305.16896, 2023.
\newblock URL \url{https://api.semanticscholar.org/CorpusID:258947061}.

\bibitem[Jaech et~al.(2024)Jaech, Kalai, Lerer, Richardson, El-Kishky, Low, Helyar, Madry, Beutel, Carney, et~al.]{o1}
Aaron Jaech, Adam Kalai, Adam Lerer, Adam Richardson, Ahmed El-Kishky, Aiden Low, Alec Helyar, Aleksander Madry, Alex Beutel, Alex Carney, et~al.
\newblock Openai o1 system card.
\newblock \emph{arXiv preprint arXiv:2412.16720}, 2024.

\bibitem[Jain et~al.(2024)Jain, Han, Gu, Li, Yan, Zhang, Wang, Solar-Lezama, Sen, and Stoica]{jain2024livecodebench}
Naman Jain, King Han, Alex Gu, Wen-Ding Li, Fanjia Yan, Tianjun Zhang, Sida Wang, Armando Solar-Lezama, Koushik Sen, and Ion Stoica.
\newblock Livecodebench: Holistic and contamination free evaluation of large language models for code.
\newblock \emph{arXiv preprint arXiv:2403.07974}, 2024.

\bibitem[Jain et~al.(2025)Jain, Singh, Shetty, Zheng, Sen, and Stoica]{Jain2025R2EGymPE}
Naman Jain, Jaskirat Singh, Manish Shetty, Liang Zheng, Koushik Sen, and Ion Stoica.
\newblock R2e-gym: Procedural environments and hybrid verifiers for scaling open-weights swe agents.
\newblock \emph{ArXiv}, abs/2504.07164, 2025.
\newblock URL \url{https://api.semanticscholar.org/CorpusID:277667306}.

\bibitem[Jiang et~al.(2025)Jiang, Lin, Cao, Tian, Kang, Wang, Sun, and Han]{Jiang2025DeepRetrievalHR}
Pengcheng Jiang, Jiacheng Lin, Lang Cao, Runchu Tian, SeongKu Kang, Zifeng Wang, Jimeng Sun, and Jiawei Han.
\newblock Deepretrieval: Hacking real search engines and retrievers with large language models via reinforcement learning.
\newblock \emph{ArXiv}, abs/2503.00223, 2025.
\newblock URL \url{https://api.semanticscholar.org/CorpusID:276742133}.

\bibitem[Jiang et~al.(2023)Jiang, Xu, Gao, Sun, Liu, Dwivedi-Yu, Yang, Callan, and Neubig]{Jiang2023ActiveRA}
Zhengbao Jiang, Frank~F. Xu, Luyu Gao, Zhiqing Sun, Qian Liu, Jane Dwivedi-Yu, Yiming Yang, Jamie Callan, and Graham Neubig.
\newblock Active retrieval augmented generation.
\newblock \emph{ArXiv}, abs/2305.06983, 2023.
\newblock URL \url{https://api.semanticscholar.org/CorpusID:258615731}.

\bibitem[Jin et~al.(2024)Jin, Yoon, Han, and Arik]{Jin2024LongContextLM}
Bowen Jin, Jinsung Yoon, Jiawei Han, and Sercan~{\"O}. Arik.
\newblock Long-context llms meet rag: Overcoming challenges for long inputs in rag.
\newblock \emph{ArXiv}, abs/2410.05983, 2024.
\newblock URL \url{https://api.semanticscholar.org/CorpusID:273229050}.

\bibitem[Jin et~al.(2025)Jin, Zeng, Yue, Wang, Zamani, and Han]{searchr1}
Bowen Jin, Hansi Zeng, Zhenrui Yue, Dong Wang, Hamed Zamani, and Jiawei Han.
\newblock Search-r1: Training llms to reason and leverage search engines with reinforcement learning.
\newblock \emph{ArXiv}, abs/2503.09516, 2025.
\newblock URL \url{https://api.semanticscholar.org/CorpusID:276937772}.

\bibitem[Joshi et~al.(2017)Joshi, Choi, Weld, and Zettlemoyer]{joshi2017triviaqa}
Mandar Joshi, Eunsol Choi, Daniel~S Weld, and Luke Zettlemoyer.
\newblock Triviaqa: A large scale distantly supervised challenge dataset for reading comprehension.
\newblock \emph{arXiv preprint arXiv:1705.03551}, 2017.

\bibitem[Karpukhin et~al.(2020)Karpukhin, Oğuz, Min, Lewis, Wu, Edunov, Chen, and tau Yih]{Karpukhin2020DensePR}
Vladimir Karpukhin, Barlas Oğuz, Sewon Min, Patrick Lewis, Ledell~Yu Wu, Sergey Edunov, Danqi Chen, and Wen tau Yih.
\newblock Dense passage retrieval for open-domain question answering.
\newblock \emph{ArXiv}, abs/2004.04906, 2020.
\newblock URL \url{https://api.semanticscholar.org/CorpusID:215737187}.

\bibitem[Ke et~al.(2025)Ke, Jiao, Ming, Nguyen, Xu, Long, Li, Qin, Wang, Savarese, Xiong, and Joty]{Ke2025ASO}
Zixuan Ke, Fangkai Jiao, Yifei Ming, Xuan-Phi Nguyen, Austin Xu, Do~Xuan Long, Minzhi Li, Chengwei Qin, PeiFeng Wang, Silvio Savarese, Caiming Xiong, and Shafiq Joty.
\newblock A survey of frontiers in llm reasoning: Inference scaling, learning to reason, and agentic systems.
\newblock \emph{Trans. Mach. Learn. Res.}, 2025, 2025.
\newblock URL \url{https://api.semanticscholar.org/CorpusID:277781085}.

\bibitem[Kim et~al.(2023)Kim, Baldi, and McAleer]{Kim2023LanguageMC}
Geunwoo Kim, Pierre Baldi, and Stephen~Marcus McAleer.
\newblock Language models can solve computer tasks.
\newblock \emph{ArXiv}, abs/2303.17491, 2023.
\newblock URL \url{https://api.semanticscholar.org/CorpusID:257834038}.

\bibitem[Kong et~al.(2023)Kong, Ruan, Chen, Zhang, Bao, Shi, Du, Hu, Mao, Li, Zeng, and Zhao]{Kong2023TPTUv2BT}
Yilun Kong, Jingqing Ruan, Yihong Chen, Bin Zhang, Tianpeng Bao, Shiwei Shi, Guoqing Du, Xiaoru Hu, Hangyu Mao, Ziyue Li, Xingyu Zeng, and Rui Zhao.
\newblock Tptu-v2: Boosting task planning and tool usage of large language model-based agents in real-world systems.
\newblock \emph{ArXiv}, abs/2311.11315, 2023.
\newblock URL \url{https://api.semanticscholar.org/CorpusID:265294410}.

\bibitem[Kwiatkowski et~al.(2019)Kwiatkowski, Palomaki, Redfield, Collins, Parikh, Alberti, Epstein, Polosukhin, Devlin, Lee, et~al.]{kwiatkowski2019natural}
Tom Kwiatkowski, Jennimaria Palomaki, Olivia Redfield, Michael Collins, Ankur Parikh, Chris Alberti, Danielle Epstein, Illia Polosukhin, Jacob Devlin, Kenton Lee, et~al.
\newblock Natural questions: a benchmark for question answering research.
\newblock \emph{Transactions of the Association for Computational Linguistics}, 7:\penalty0 453--466, 2019.

\bibitem[Lewis et~al.(2020)Lewis, Perez, Piktus, Petroni, Karpukhin, Goyal, Kuttler, Lewis, tau Yih, Rockt{\"a}schel, Riedel, and Kiela]{Lewis2020RetrievalAugmentedGF}
Patrick Lewis, Ethan Perez, Aleksandara Piktus, Fabio Petroni, Vladimir Karpukhin, Naman Goyal, Heinrich Kuttler, Mike Lewis, Wen tau Yih, Tim Rockt{\"a}schel, Sebastian Riedel, and Douwe Kiela.
\newblock Retrieval-augmented generation for knowledge-intensive nlp tasks.
\newblock \emph{ArXiv}, abs/2005.11401, 2020.
\newblock URL \url{https://api.semanticscholar.org/CorpusID:218869575}.

\bibitem[Lewkowycz et~al.(2022)Lewkowycz, Andreassen, Dohan, Dyer, Michalewski, Ramasesh, Slone, Anil, Schlag, Gutman-Solo, Wu, Neyshabur, Gur-Ari, and Misra]{lewkowycz2022solving}
Aitor Lewkowycz, Anders~Johan Andreassen, David Dohan, Ethan Dyer, Henryk Michalewski, Vinay~Venkatesh Ramasesh, Ambrose Slone, Cem Anil, Imanol Schlag, Theo Gutman-Solo, Yuhuai Wu, Behnam Neyshabur, Guy Gur-Ari, and Vedant Misra.
\newblock Solving quantitative reasoning problems with language models.
\newblock In Alice~H. Oh, Alekh Agarwal, Danielle Belgrave, and Kyunghyun Cho (eds.), \emph{Advances in Neural Information Processing Systems}, 2022.
\newblock URL \url{https://openreview.net/forum?id=IFXTZERXdM7}.

\bibitem[Li et~al.(2024)Li, Zhang, Liu, Fan, Zhang, Zhu, Wei, Pan, Li, and Chen]{Li2024CodeSTB}
Haoyang Li, Jing Zhang, Hanbing Liu, Ju~Fan, Xiaokang Zhang, Jun Zhu, Renjie Wei, Hongyan Pan, Cuiping Li, and Hong Chen.
\newblock Codes: Towards building open-source language models for text-to-sql.
\newblock \emph{Proceedings of the ACM on Management of Data}, 2:\penalty0 1 -- 28, 2024.
\newblock URL \url{https://api.semanticscholar.org/CorpusID:267938784}.

\bibitem[Li et~al.(2025{\natexlab{a}})Li, Wu, Zhang, Huang, Zhang, Jiang, Wang, Zhang, Chen, Shi, Chen, and Li]{Li2025OmniSQLSH}
Haoyang Li, Shang Wu, Xiaokang Zhang, Xinmei Huang, Jing Zhang, Fuxin Jiang, Shuai Wang, Tieying Zhang, Jianjun Chen, Rui Shi, Hong Chen, and Cuiping Li.
\newblock Omnisql: Synthesizing high-quality text-to-sql data at scale.
\newblock \emph{ArXiv}, abs/2503.02240, 2025{\natexlab{a}}.
\newblock URL \url{https://api.semanticscholar.org/CorpusID:276774742}.

\bibitem[Li et~al.(2023)Li, Hui, Cheng, Qin, Ma, Huo, Huang, Du, Si, and Li]{Li2023GraphixT5MP}
Jinyang Li, Binyuan Hui, Reynold Cheng, Bowen Qin, Chenhao Ma, Nan Huo, Fei Huang, Wenyu Du, Luo Si, and Yongbin Li.
\newblock Graphix-t5: Mixing pre-trained transformers with graph-aware layers for text-to-sql parsing.
\newblock \emph{ArXiv}, abs/2301.07507, 2023.
\newblock URL \url{https://api.semanticscholar.org/CorpusID:255998567}.

\bibitem[Li et~al.(2025{\natexlab{b}})Li, Zhang, Yin, Zhang, Ou, Wu, Yin, Li, Tao, Wang, Shen, Zhang, Zhang, Wu, Jiang, Yan, Xie, Huang, and Zhou]{Li2025WebSailorNS}
Kuan Li, Zhongwang Zhang, Huifeng Yin, Liwen Zhang, Litu Ou, Jialong Wu, Wenbiao Yin, Baixuan Li, Zhengwei Tao, Xinyu Wang, Weizhou Shen, Junkai Zhang, Dingchu Zhang, Xixi Wu, Yong Jiang, Ming Yan, Pengjun Xie, Fei Huang, and Jingren Zhou.
\newblock Websailor: Navigating super-human reasoning for web agent.
\newblock \emph{ArXiv}, abs/2507.02592, 2025{\natexlab{b}}.
\newblock URL \url{https://api.semanticscholar.org/CorpusID:280078605}.

\bibitem[Li et~al.(2025{\natexlab{c}})Li, Zou, and Liu]{Li2025ToRLST}
Xuefeng Li, Haoyang Zou, and Pengfei Liu.
\newblock Torl: Scaling tool-integrated rl, 2025{\natexlab{c}}.
\newblock URL \url{https://arxiv.org/abs/2503.23383}.

\bibitem[Liu et~al.(2025)Liu, Hegde, Cao, Zhu, Li, Griggs, Tang, Malik, Hakhamaneshi, Liaw, Moritz, Zaharia, Gonzalez, and Stoica]{liu2025skyrlsql}
Shu Liu, Sumanth Hegde, Shiyi Cao, Alan Zhu, Dacheng Li, Tyler Griggs, Eric Tang, Akshay Malik, Kourosh Hakhamaneshi, Richard Liaw, Philipp Moritz, Matei Zaharia, Joseph~E. Gonzalez, and Ion Stoica.
\newblock Skyrl-sql: Matching gpt-4o and o4-mini on text2sql with multi-turn rl, 2025.
\newblock Notion Blog.

\bibitem[Liu et~al.(2024)Liu, Shen, Li, Ma, Jiang, xin Zhang, Fan, Li, Tang, and Luo]{Liu2024ASO}
Xinyu Liu, Shuyu Shen, Boyan Li, Peixian Ma, Runzhi Jiang, Yu~xin Zhang, Ju~Fan, Guoliang Li, Nan Tang, and Yuyu Luo.
\newblock A survey of text-to-sql in the era of llms: Where are we, and where are we going?
\newblock \emph{IEEE Transactions on Knowledge and Data Engineering}, 2024.
\newblock URL \url{https://api.semanticscholar.org/CorpusID:271843296}.

\bibitem[Lu et~al.(2023)Lu, Peng, Cheng, Galley, Chang, Wu, Zhu, and Gao]{Lu2023ChameleonPC}
Pan Lu, Baolin Peng, Hao Cheng, Michel Galley, Kai-Wei Chang, Ying~Nian Wu, Song-Chun Zhu, and Jianfeng Gao.
\newblock Chameleon: Plug-and-play compositional reasoning with large language models.
\newblock \emph{ArXiv}, abs/2304.09842, 2023.
\newblock URL \url{https://api.semanticscholar.org/CorpusID:258212542}.

\bibitem[Ma et~al.(2025)Ma, Liu, Jiang, Zhang, Ma, and Chen]{Ma2025GeneralReasonerAL}
Xueguang Ma, Qian Liu, Dongfu Jiang, Ge~Zhang, Zejun Ma, and Wenhu Chen.
\newblock General-reasoner: Advancing llm reasoning across all domains.
\newblock \emph{ArXiv}, abs/2505.14652, 2025.
\newblock URL \url{https://api.semanticscholar.org/CorpusID:278768680}.

\bibitem[Mai et~al.(2025)Mai, Xu, Wang, Hu, Zhang, Zhang, et~al.]{mai2025agent}
Xinji Mai, Haotian Xu, Weinong Wang, Jian Hu, Yingying Zhang, Wenqiang Zhang, et~al.
\newblock Agent rl scaling law: Agent rl with spontaneous code execution for mathematical problem solving.
\newblock \emph{arXiv preprint arXiv:2505.07773}, 2025.

\bibitem[Mallen et~al.(2022)Mallen, Asai, Zhong, Das, Hajishirzi, and Khashabi]{mallen2022not}
Alex Mallen, Akari Asai, Victor Zhong, Rajarshi Das, Hannaneh Hajishirzi, and Daniel Khashabi.
\newblock When not to trust language models: Investigating effectiveness and limitations of parametric and non-parametric memories.
\newblock \emph{arXiv preprint arXiv:2212.10511}, 7, 2022.

\bibitem[Mialon et~al.(2023)Mialon, Fourrier, Wolf, LeCun, and Scialom]{mialon2023gaia}
Gr{\'e}goire Mialon, Cl{\'e}mentine Fourrier, Thomas Wolf, Yann LeCun, and Thomas Scialom.
\newblock Gaia: a benchmark for general ai assistants.
\newblock In \emph{The Twelfth International Conference on Learning Representations}, 2023.

\bibitem[Moritz et~al.(2017)Moritz, Nishihara, Wang, Tumanov, Liaw, Liang, Paul, Jordan, and Stoica]{Moritz2017RayAD}
Philipp Moritz, Robert Nishihara, Stephanie Wang, Alexey Tumanov, Richard Liaw, Eric Liang, William Paul, Michael~I. Jordan, and Ion Stoica.
\newblock Ray: A distributed framework for emerging ai applications.
\newblock \emph{ArXiv}, abs/1712.05889, 2017.
\newblock URL \url{https://api.semanticscholar.org/CorpusID:34552495}.

\bibitem[Moshkov et~al.(2025)Moshkov, Hanley, Sorokin, Toshniwal, Henkel, Schifferer, Du, and Gitman]{Moshkov2025AIMO2WS}
Ivan Moshkov, Darragh Hanley, Ivan Sorokin, Shubham Toshniwal, Christof Henkel, Benedikt Schifferer, Wei Du, and Igor Gitman.
\newblock Aimo-2 winning solution: Building state-of-the-art mathematical reasoning models with openmathreasoning dataset, 2025.
\newblock URL \url{https://arxiv.org/abs/2504.16891}.

\bibitem[{OpenAI}(2025)]{openai2024gpt41}
{OpenAI}.
\newblock Introducing gpt-4.1 in the api.
\newblock https://openai.com/index/gpt-4-1/, 2025.

\bibitem[Papicchio et~al.(2025)Papicchio, Rossi, Cagliero, and Papotti]{Papicchio2025Think2SQLRL}
Simone Papicchio, Simone Rossi, Luca Cagliero, and Paolo Papotti.
\newblock Think2sql: Reinforce llm reasoning capabilities for text2sql.
\newblock \emph{ArXiv}, abs/2504.15077, 2025.
\newblock URL \url{https://api.semanticscholar.org/CorpusID:277955819}.

\bibitem[Peng et~al.(2023)Peng, Yang, Chen, Smith, Pournejatian, Costa, Martin, Flores, Zhang, Magoc, Lipori, Mitchell, Ospina, Ahmed, Hogan, Shenkman, Guo, Bian, and Wu]{Peng2023ASO}
C.A.I. Peng, Xi~Yang, Aokun Chen, Kaleb~E. Smith, Nima~M. Pournejatian, Anthony~B Costa, Cheryl Martin, Mona~G. Flores, Ying Zhang, Tanja Magoc, Gloria~P. Lipori, Duane~A. Mitchell, Naykky M.~Singh Ospina, Mustafa~Mamon Ahmed, William~R. Hogan, Elizabeth~A. Shenkman, Yi~Guo, Jiang Bian, and Yonghui Wu.
\newblock A study of generative large language model for medical research and healthcare.
\newblock \emph{NPJ Digital Medicine}, 6, 2023.
\newblock URL \url{https://api.semanticscholar.org/CorpusID:258841310}.

\bibitem[Phan et~al.(2025)Phan, Gatti, Han, Li, Hu, Zhang, Zhang, Shaaban, Ling, Shi, et~al.]{phan2025humanity}
Long Phan, Alice Gatti, Ziwen Han, Nathaniel Li, Josephina Hu, Hugh Zhang, Chen Bo~Calvin Zhang, Mohamed Shaaban, John Ling, Sean Shi, et~al.
\newblock Humanity's last exam.
\newblock \emph{arXiv preprint arXiv:2501.14249}, 2025.

\bibitem[Plaat et~al.(2025)Plaat, van Duijn, van Stein, Preuss, van~der Putten, and Batenburg]{Plaat2025AgenticLL}
Aske Plaat, Max~J. van Duijn, Niki van Stein, Mike Preuss, Peter van~der Putten, and Kees~Joost Batenburg.
\newblock Agentic large language models, a survey.
\newblock \emph{ArXiv}, abs/2503.23037, 2025.
\newblock URL \url{https://api.semanticscholar.org/CorpusID:277451794}.

\bibitem[Pourreza et~al.(2025)Pourreza, Talaei, Sun, Wan, Li, Mirhoseini, Saberi, and Arik]{Pourreza2025ReasoningSQLRL}
Mohammadreza Pourreza, Shayan Talaei, Ruoxi Sun, Xingchen Wan, Hailong Li, Azalia Mirhoseini, Amin Saberi, and Sercan~{\"O}. Arik.
\newblock Reasoning-sql: Reinforcement learning with sql tailored partial rewards for reasoning-enhanced text-to-sql.
\newblock \emph{ArXiv}, abs/2503.23157, 2025.
\newblock URL \url{https://api.semanticscholar.org/CorpusID:277452634}.

\bibitem[Press et~al.(2022)Press, Zhang, Min, Schmidt, Smith, and Lewis]{press2022measuring}
Ofir Press, Muru Zhang, Sewon Min, Ludwig Schmidt, Noah~A Smith, and Mike Lewis.
\newblock Measuring and narrowing the compositionality gap in language models.
\newblock \emph{arXiv preprint arXiv:2210.03350}, 2022.

\bibitem[Qian et~al.(2025)Qian, Acikgoz, He, Wang, Chen, Hakkani-T{\"u}r, Tur, and Ji]{Qian2025ToolRLRI}
Cheng Qian, Emre~Can Acikgoz, Qi~He, Hongru Wang, Xiusi Chen, Dilek Hakkani-T{\"u}r, Gokhan Tur, and Heng Ji.
\newblock Toolrl: Reward is all tool learning needs.
\newblock \emph{arXiv preprint arXiv:2504.13958}, 2025.

\bibitem[Qu et~al.(2024)Qu, Dai, Wei, Cai, Wang, Yin, Xu, and Wen]{Qu2024ToolLW}
Changle Qu, Sunhao Dai, Xiaochi Wei, Hengyi Cai, Shuaiqiang Wang, Dawei Yin, Jun Xu, and Jirong Wen.
\newblock Tool learning with large language models: A survey.
\newblock \emph{ArXiv}, abs/2405.17935, 2024.
\newblock URL \url{https://api.semanticscholar.org/CorpusID:270067624}.

\bibitem[Quan et~al.(2025)Quan, Yang, Yu, Zheng, Liu, Yang, Ren, Gao, Miao, Feng, Wang, Yang, Cui, Fan, Zhang, Hui, and Lin]{Quan2025CodeEloBC}
Shanghaoran Quan, Jiaxin Yang, Bowen Yu, Bo~Zheng, Dayiheng Liu, An~Yang, Xuancheng Ren, Bofei Gao, Yibo Miao, Yunlong Feng, Zekun Wang, Jian Yang, Zeyu Cui, Yang Fan, Yichang Zhang, Binyuan Hui, and Junyang Lin.
\newblock Codeelo: Benchmarking competition-level code generation of llms with human-comparable elo ratings.
\newblock \emph{ArXiv}, abs/2501.01257, 2025.
\newblock URL \url{https://api.semanticscholar.org/CorpusID:275212089}.

\bibitem[Qwen et~al.(2025)Qwen, :, Yang, Yang, Zhang, Hui, Zheng, Yu, Li, Liu, Huang, Wei, Lin, Yang, Tu, Zhang, Yang, Yang, Zhou, Lin, Dang, Lu, Bao, Yang, Yu, Li, Xue, Zhang, Zhu, Men, Lin, Li, Tang, Xia, Ren, Ren, Fan, Su, Zhang, Wan, Liu, Cui, Zhang, and Qiu]{qwen2025qwen25technicalreport}
Qwen, :, An~Yang, Baosong Yang, Beichen Zhang, Binyuan Hui, Bo~Zheng, Bowen Yu, Chengyuan Li, Dayiheng Liu, Fei Huang, Haoran Wei, Huan Lin, Jian Yang, Jianhong Tu, Jianwei Zhang, Jianxin Yang, Jiaxi Yang, Jingren Zhou, Junyang Lin, Kai Dang, Keming Lu, Keqin Bao, Kexin Yang, Le~Yu, Mei Li, Mingfeng Xue, Pei Zhang, Qin Zhu, Rui Men, Runji Lin, Tianhao Li, Tianyi Tang, Tingyu Xia, Xingzhang Ren, Xuancheng Ren, Yang Fan, Yang Su, Yichang Zhang, Yu~Wan, Yuqiong Liu, Zeyu Cui, Zhenru Zhang, and Zihan Qiu.
\newblock Qwen2.5 technical report, 2025.
\newblock URL \url{https://arxiv.org/abs/2412.15115}.

\bibitem[Rafailov et~al.(2023)Rafailov, Sharma, Mitchell, Ermon, Manning, and Finn]{Rafailov2023DirectPO}
Rafael Rafailov, Archit Sharma, Eric Mitchell, Stefano Ermon, Christopher~D. Manning, and Chelsea Finn.
\newblock Direct preference optimization: Your language model is secretly a reward model.
\newblock \emph{ArXiv}, abs/2305.18290, 2023.
\newblock URL \url{https://api.semanticscholar.org/CorpusID:258959321}.

\bibitem[Ruan et~al.(2023)Ruan, Chen, Zhang, Xu, Bao, Du, Shi, Mao, Zeng, and Zhao]{Ruan2023TPTUTP}
Jingqing Ruan, Yihong Chen, Bin Zhang, Zhiwei Xu, Tianpeng Bao, Guoqing Du, Shiwei Shi, Hangyu Mao, Xingyu Zeng, and Rui Zhao.
\newblock Tptu: Task planning and tool usage of large language model-based ai agents.
\newblock \emph{ArXiv}, abs/2308.03427, 2023.
\newblock URL \url{https://api.semanticscholar.org/CorpusID:265381326}.

\bibitem[Russell \& Norvig(2016)Russell and Norvig]{russell2016artificial}
Stuart~J Russell and Peter Norvig.
\newblock \emph{Artificial intelligence: a modern approach}.
\newblock pearson, 2016.

\bibitem[Schick et~al.(2023)Schick, Dwivedi-Yu, Dess{\`i}, Raileanu, Lomeli, Zettlemoyer, Cancedda, and Scialom]{Schick2023ToolformerLM}
Timo Schick, Jane Dwivedi-Yu, Roberto Dess{\`i}, Roberta Raileanu, Maria Lomeli, Luke Zettlemoyer, Nicola Cancedda, and Thomas Scialom.
\newblock Toolformer: Language models can teach themselves to use tools.
\newblock \emph{ArXiv}, abs/2302.04761, 2023.
\newblock URL \url{https://api.semanticscholar.org/CorpusID:256697342}.

\bibitem[Schulman et~al.(2017)Schulman, Wolski, Dhariwal, Radford, and Klimov]{Schulman2017ProximalPO}
John Schulman, Filip Wolski, Prafulla Dhariwal, Alec Radford, and Oleg Klimov.
\newblock Proximal policy optimization algorithms.
\newblock \emph{ArXiv}, abs/1707.06347, 2017.
\newblock URL \url{https://api.semanticscholar.org/CorpusID:28695052}.

\bibitem[Shao et~al.(2024)Shao, Wang, Zhu, Xu, Song, Zhang, Li, Wu, and Guo]{Shao2024DeepSeekMathPT}
Zhihong Shao, Peiyi Wang, Qihao Zhu, Runxin Xu, Jun-Mei Song, Mingchuan Zhang, Y.~K. Li, Yu~Wu, and Daya Guo.
\newblock Deepseekmath: Pushing the limits of mathematical reasoning in open language models.
\newblock \emph{ArXiv}, abs/2402.03300, 2024.
\newblock URL \url{https://api.semanticscholar.org/CorpusID:267412607}.

\bibitem[Shen et~al.(2023)Shen, Song, Tan, Li, Lu, and Zhuang]{Shen2023HuggingGPTSA}
Yongliang Shen, Kaitao Song, Xu~Tan, Dongsheng Li, Weiming Lu, and Yue~Ting Zhuang.
\newblock Hugginggpt: Solving ai tasks with chatgpt and its friends in hugging face.
\newblock \emph{ArXiv}, abs/2303.17580, 2023.
\newblock URL \url{https://api.semanticscholar.org/CorpusID:257833781}.

\bibitem[Shen(2024)]{Shen2024LLMWT}
Zhuocheng Shen.
\newblock Llm with tools: A survey.
\newblock \emph{ArXiv}, abs/2409.18807, 2024.
\newblock URL \url{https://api.semanticscholar.org/CorpusID:272968969}.

\bibitem[Sheng et~al.(2024)Sheng, Zhang, Ye, Wu, Zhang, Zhang, Peng, Lin, and Wu]{sheng2024hybridflow}
Guangming Sheng, Chi Zhang, Zilingfeng Ye, Xibin Wu, Wang Zhang, Ru~Zhang, Yanghua Peng, Haibin Lin, and Chuan Wu.
\newblock Hybridflow: A flexible and efficient rlhf framework.
\newblock \emph{arXiv preprint arXiv: 2409.19256}, 2024.

\bibitem[Song et~al.(2025)Song, Jiang, Min, Chen, Chen, Zhao, Fang, and Wen]{song2025r1searcherincentivizingsearchcapability}
Huatong Song, Jinhao Jiang, Yingqian Min, Jie Chen, Zhipeng Chen, Wayne~Xin Zhao, Lei Fang, and Ji-Rong Wen.
\newblock R1-searcher: Incentivizing the search capability in llms via reinforcement learning, 2025.
\newblock URL \url{https://arxiv.org/abs/2503.05592}.

\bibitem[Su et~al.(2025)Su, Wang, Ren, Lin, and Chen]{pixelreasoner}
Alex Su, Haozhe Wang, Weiming Ren, Fangzhen Lin, and Wenhu Chen.
\newblock Pixel reasoner: Incentivizing pixel-space reasoning with curiosity-driven reinforcement learning.
\newblock \emph{ArXiv}, abs/2505.15966, 2025.
\newblock URL \url{https://api.semanticscholar.org/CorpusID:278789415}.

\bibitem[Sun et~al.(2025)Sun, Song, Wang, Ren, Jiang, Zhang, Bai, Deng, Zhao, Liu, et~al.]{sun2025simpledeepsearcher}
Shuang Sun, Huatong Song, Yuhao Wang, Ruiyang Ren, Jinhao Jiang, Junjie Zhang, Fei Bai, Jia Deng, Wayne~Xin Zhao, Zheng Liu, et~al.
\newblock Simpledeepsearcher: Deep information seeking via web-powered reasoning trajectory synthesis.
\newblock \emph{arXiv preprint arXiv:2505.16834}, 2025.

\bibitem[Team et~al.(2024)Team, Georgiev, Lei, Burnell, Bai, Gulati, Tanzer, Vincent, Pan, Wang, et~al.]{gemini}
Gemini Team, Petko Georgiev, Ving~Ian Lei, Ryan Burnell, Libin Bai, Anmol Gulati, Garrett Tanzer, Damien Vincent, Zhufeng Pan, Shibo Wang, et~al.
\newblock Gemini 1.5: Unlocking multimodal understanding across millions of tokens of context.
\newblock \emph{arXiv preprint arXiv:2403.05530}, 2024.

\bibitem[{THUDM}(2024)]{slime2024}
{THUDM}.
\newblock slime: A llm post-training framework aiming at scaling rl.
\newblock \url{https://github.com/THUDM/slime}, 2024.
\newblock Software framework for LLM post-training with reinforcement learning scaling.

\bibitem[Trivedi et~al.(2022{\natexlab{a}})Trivedi, Balasubramanian, Khot, and Sabharwal]{Trivedi2022InterleavingRW}
H.~Trivedi, Niranjan Balasubramanian, Tushar Khot, and Ashish Sabharwal.
\newblock Interleaving retrieval with chain-of-thought reasoning for knowledge-intensive multi-step questions.
\newblock \emph{ArXiv}, abs/2212.10509, 2022{\natexlab{a}}.
\newblock URL \url{https://api.semanticscholar.org/CorpusID:254877499}.

\bibitem[Trivedi et~al.(2022{\natexlab{b}})Trivedi, Balasubramanian, Khot, and Sabharwal]{trivedi2022musique}
Harsh Trivedi, Niranjan Balasubramanian, Tushar Khot, and Ashish Sabharwal.
\newblock Musique: Multihop questions via single-hop question composition.
\newblock \emph{Transactions of the Association for Computational Linguistics}, 10:\penalty0 539--554, 2022{\natexlab{b}}.

\bibitem[Wang et~al.(2019)Wang, Shin, Liu, Polozov, and Richardson]{Wang2019RATSQLRS}
Bailin Wang, Richard Shin, Xiaodong Liu, Oleksandr Polozov, and Matthew Richardson.
\newblock Rat-sql: Relation-aware schema encoding and linking for text-to-sql parsers.
\newblock In \emph{Annual Meeting of the Association for Computational Linguistics}, 2019.
\newblock URL \url{https://api.semanticscholar.org/CorpusID:207863446}.

\bibitem[Wang et~al.(2025{\natexlab{a}})Wang, Li, Qu, Zhu, Xu, Chu, and Lin]{wang2025code}
Haozhe Wang, Long Li, Chao Qu, Fengming Zhu, Weidi Xu, Wei Chu, and Fangzhen Lin.
\newblock To code or not to code? adaptive tool integration for math language models via expectation-maximization.
\newblock \emph{arXiv preprint arXiv:2502.00691}, 2025{\natexlab{a}}.

\bibitem[Wang et~al.(2024{\natexlab{a}})Wang, Qin, Lin, Pan, and Wong]{10.1145/3626772.3661381}
Hongru Wang, Yujia Qin, Yankai Lin, Jeff~Z. Pan, and Kam-Fai Wong.
\newblock Empowering large language models: Tool learning for real-world interaction.
\newblock In \emph{Proceedings of the 47th International ACM SIGIR Conference on Research and Development in Information Retrieval}, SIGIR '24, pp.\  2983–2986, New York, NY, USA, 2024{\natexlab{a}}. Association for Computing Machinery.
\newblock ISBN 9798400704314.
\newblock \doi{10.1145/3626772.3661381}.
\newblock URL \url{https://doi.org/10.1145/3626772.3661381}.

\bibitem[Wang et~al.(2023)Wang, Ren, Zhou, Lu, Luo, Shi, Zhang, Song, Zhan, and Li]{wang2023mathcoderseamlesscodeintegration}
Ke~Wang, Houxing Ren, Aojun Zhou, Zimu Lu, Sichun Luo, Weikang Shi, Renrui Zhang, Linqi Song, Mingjie Zhan, and Hongsheng Li.
\newblock Mathcoder: Seamless code integration in llms for enhanced mathematical reasoning, 2023.
\newblock URL \url{https://arxiv.org/abs/2310.03731}.

\bibitem[Wang et~al.(2022)Wang, Yang, Huang, Jiao, Yang, Jiang, Majumder, and Wei]{Wang2022TextEB}
Liang Wang, Nan Yang, Xiaolong Huang, Binxing Jiao, Linjun Yang, Daxin Jiang, Rangan Majumder, and Furu Wei.
\newblock Text embeddings by weakly-supervised contrastive pre-training.
\newblock \emph{ArXiv}, abs/2212.03533, 2022.
\newblock URL \url{https://api.semanticscholar.org/CorpusID:254366618}.

\bibitem[Wang et~al.(2025{\natexlab{b}})Wang, Xiong, Chen, Gao, Guo, He, Huang, Liu, Li, Li, et~al.]{wang2025reinforcement}
Weixun Wang, Shaopan Xiong, Gengru Chen, Wei Gao, Sheng Guo, Yancheng He, Ju~Huang, Jiaheng Liu, Zhendong Li, Xiaoyang Li, et~al.
\newblock Reinforcement learning optimization for large-scale learning: An efficient and user-friendly scaling library.
\newblock \emph{arXiv preprint arXiv:2506.06122}, 2025{\natexlab{b}}.

\bibitem[Wang et~al.(2024{\natexlab{b}})Wang, Li, Song, Xu, Tang, Zhuge, Pan, Song, Li, Singh, Tran, Li, Ma, Zheng, Qian, Shao, Muennighoff, Zhang, Hui, Lin, Brennan, Peng, Ji, and Neubig]{Wang2024OpenHandsAO}
Xingyao Wang, Boxuan Li, Yufan Song, Frank~F. Xu, Xiangru Tang, Mingchen Zhuge, Jiayi Pan, Yueqi Song, Bowen Li, Jaskirat Singh, Hoang~H. Tran, Fuqiang Li, Ren Ma, Mingzhang Zheng, Bill Qian, Yanjun Shao, Niklas Muennighoff, Yizhe Zhang, Binyuan Hui, Junyang Lin, Robert Brennan, Hao Peng, Heng Ji, and Graham Neubig.
\newblock Openhands: An open platform for ai software developers as generalist agents.
\newblock In \emph{International Conference on Learning Representations}, 2024{\natexlab{b}}.
\newblock URL \url{https://api.semanticscholar.org/CorpusID:271404773}.

\bibitem[Wang et~al.(2025{\natexlab{c}})Wang, Wang, Wang, Zhang, Li, Yang, Yu, Nguyen, Liu, Gottlieb, Lam, Lu, Cho, Wu, Li, Wang, Choi, and Li]{Wang2025RAGENUS}
Zihan Wang, Kangrui Wang, Qineng Wang, Pingyue Zhang, Linjie Li, Zhengyuan Yang, Kefan Yu, Minh~Nhat Nguyen, Licheng Liu, Eli Gottlieb, Monica Lam, Yiping Lu, Kyunghyun Cho, Jiajun Wu, Fei-Fei Li, Lijuan Wang, Yejin Choi, and Manling Li.
\newblock Ragen: Understanding self-evolution in llm agents via multi-turn reinforcement learning.
\newblock \emph{ArXiv}, abs/2504.20073, 2025{\natexlab{c}}.
\newblock URL \url{https://api.semanticscholar.org/CorpusID:278170861}.

\bibitem[Wei et~al.(2022)Wei, Wang, Schuurmans, Bosma, Chi, Xia, Le, and Zhou]{Wei2022ChainOT}
Jason Wei, Xuezhi Wang, Dale Schuurmans, Maarten Bosma, Ed~H. Chi, F.~Xia, Quoc Le, and Denny Zhou.
\newblock Chain of thought prompting elicits reasoning in large language models.
\newblock \emph{ArXiv}, abs/2201.11903, 2022.
\newblock URL \url{https://api.semanticscholar.org/CorpusID:246411621}.

\bibitem[Wooldridge(1999)]{wooldridge1999intelligent}
Michael Wooldridge.
\newblock Intelligent agents.
\newblock \emph{Multiagent systems: A modern approach to distributed artificial intelligence}, 1:\penalty0 27--73, 1999.

\bibitem[Wu \& Xie(2024)Wu and Xie]{wu2024v}
Penghao Wu and Saining Xie.
\newblock V?: Guided visual search as a core mechanism in multimodal llms.
\newblock In \emph{Proceedings of the IEEE/CVF Conference on Computer Vision and Pattern Recognition}, pp.\  13084--13094, 2024.

\bibitem[Yang et~al.(2023)Yang, Song, Li, Zhao, Ge, Li, and Shan]{Yang2023GPT4ToolsTL}
Rui Yang, Lin Song, Yanwei Li, Sijie Zhao, Yixiao Ge, Xiu Li, and Ying Shan.
\newblock Gpt4tools: Teaching large language model to use tools via self-instruction.
\newblock \emph{ArXiv}, abs/2305.18752, 2023.
\newblock URL \url{https://api.semanticscholar.org/CorpusID:258967184}.

\bibitem[Yang et~al.(2018)Yang, Qi, Zhang, Bengio, Cohen, Salakhutdinov, and Manning]{yang2018hotpotqa}
Zhilin Yang, Peng Qi, Saizheng Zhang, Yoshua Bengio, William~W Cohen, Ruslan Salakhutdinov, and Christopher~D Manning.
\newblock Hotpotqa: A dataset for diverse, explainable multi-hop question answering.
\newblock \emph{arXiv preprint arXiv:1809.09600}, 2018.

\bibitem[Yao et~al.(2022)Yao, Zhao, Yu, Du, Shafran, Narasimhan, and Cao]{Yao2022ReActSR}
Shunyu Yao, Jeffrey Zhao, Dian Yu, Nan Du, Izhak Shafran, Karthik Narasimhan, and Yuan Cao.
\newblock React: Synergizing reasoning and acting in language models.
\newblock \emph{ArXiv}, abs/2210.03629, 2022.
\newblock URL \url{https://api.semanticscholar.org/CorpusID:252762395}.

\bibitem[Yao et~al.(2025)Yao, Liu, Chen, Chen, Fang, Hou, Li, and Chua]{Yao2025AreRM}
Zijun Yao, Yantao Liu, Yanxu Chen, Jianhui Chen, Junfeng Fang, Lei Hou, Juanzi Li, and Tat-Seng Chua.
\newblock Are reasoning models more prone to hallucination?
\newblock \emph{ArXiv}, abs/2505.23646, 2025.
\newblock URL \url{https://api.semanticscholar.org/CorpusID:278996592}.

\bibitem[Ye et~al.(2025)Ye, Huang, Xiao, Chern, Xia, and Liu]{Ye2025LIMOLI}
Yixin Ye, Zhen Huang, Yang Xiao, Ethan Chern, Shijie Xia, and Pengfei Liu.
\newblock Limo: Less is more for reasoning.
\newblock \emph{ArXiv}, abs/2502.03387, 2025.
\newblock URL \url{https://api.semanticscholar.org/CorpusID:276116748}.

\bibitem[Yu et~al.(2018)Yu, Zhang, Yang, Yasunaga, Wang, Li, Ma, Li, Yao, Roman, Zhang, and Radev]{Yu2018SpiderAL}
Tao Yu, Rui Zhang, Kai-Chou Yang, Michihiro Yasunaga, Dongxu Wang, Zifan Li, James Ma, Irene~Z Li, Qingning Yao, Shanelle Roman, Zilin Zhang, and Dragomir~R. Radev.
\newblock Spider: A large-scale human-labeled dataset for complex and cross-domain semantic parsing and text-to-sql task.
\newblock \emph{ArXiv}, abs/1809.08887, 2018.
\newblock URL \url{https://api.semanticscholar.org/CorpusID:52815560}.

\bibitem[Yue et~al.(2023)Yue, Qu, Zhang, Fu, Huang, Sun, Su, and Chen]{yue2023mammothbuildingmathgeneralist}
Xiang Yue, Xingwei Qu, Ge~Zhang, Yao Fu, Wenhao Huang, Huan Sun, Yu~Su, and Wenhu Chen.
\newblock Mammoth: Building math generalist models through hybrid instruction tuning, 2023.
\newblock URL \url{https://arxiv.org/abs/2309.05653}.

\bibitem[Zhai et~al.(2025)Zhai, Xu, He, and Yao]{Zhai2025ExCoTOR}
Bohan Zhai, Canwen Xu, Yuxiong He, and Zhewei Yao.
\newblock Excot: Optimizing reasoning for text-to-sql with execution feedback.
\newblock In \emph{Annual Meeting of the Association for Computational Linguistics}, 2025.
\newblock URL \url{https://api.semanticscholar.org/CorpusID:277321668}.

\bibitem[Zhang et~al.(2023)Zhang, Li, Cui, Cai, Liu, Fu, Huang, Zhao, Zhang, Chen, Wang, Luu, Bi, Shi, and Shi]{Zhang2023SirensSI}
Yue Zhang, Yafu Li, Leyang Cui, Deng Cai, Lemao Liu, Tingchen Fu, Xinting Huang, Enbo Zhao, Yu~Zhang, Yulong Chen, Longyue Wang, Anh~Tuan Luu, Wei Bi, Freda Shi, and Shuming Shi.
\newblock Siren's song in the ai ocean: A survey on hallucination in large language models.
\newblock \emph{ArXiv}, abs/2309.01219, 2023.
\newblock URL \url{https://api.semanticscholar.org/CorpusID:261530162}.

\end{thebibliography}
\bibliographystyle{iclr2026_conference}

\clearpage
\appendix
\section*{Appendix}
\startcontents[sections]
\printcontents[sections]{l}{1}{\setcounter{tocdepth}{3}}
\clearpage

\section{Detailed Experiment Setup}
\label{sec:exp_setup}

We evaluate \ours{} across six diverse domains to demonstrate its effectiveness in tool-augmented reasoning. Each task domain presents unique challenges and requires different tool integration strategies, allowing us to comprehensively assess the framework's adaptability and performance.

\subsection{Mathematical Reasoning with Python Executor (VT-Math)}
\label{subsec:mathtir_details}

Mathematical reasoning tasks often involve complex computations that are prone to numerical errors when performed purely through natural language reasoning. To address this limitation, we integrate a Python code interpreter tool that enables agents to execute mathematical calculations reliably and verify intermediate results. We train a mathematical-coding agent that issues Python snippets to a sandboxed interpreter and processes execution traces.

We use DeepMath~\citep{He2025DeepMath103KAL} as our training dataset. The reward function combines answer accuracy with tool usage incentives:
\begin{equation}
    R_{\text{acc}}(\mathbf{x},\mathbf{y}) = \begin{cases}
1 & \text{if } \text{match}(\mathbf{y}, \mathbf{y}_g) \\
-1 & \text{otherwise} 
\end{cases},\quad
R_{\text{tool}}(\mathbf{x},\mathbf{y}) = \begin{cases}
0 & \text{if } \text{match}(\mathbf{y}, \mathbf{y}_g) \\
-0.25 & \text{otherwise}
\end{cases}
\end{equation}
where the final reward is $R_{\text{math}} = R_{\text{acc}}(\mathbf{x}, \mathbf{y}) + R_{\text{tool}}(\mathbf{x}, \mathbf{y})$. This design encourages the model to explore Python executor usage for problem-solving while maintaining an accuracy focus.

We evaluate on multiple mathematical benchmarks: \textsc{MATH-500}~\citep{hendrycksmath2021}, \textsc{Olympiad}~\citep{he2024olympiadbench}, \textsc{Minerva}~\citep{lewkowycz2022solving}, \textsc{GSM8K}~\citep{cobbe2021gsm8k}, \textsc{AMC}, \textsc{AIME24}, and \textsc{AIME25}, using \textsc{Math-Evaluation-Harness}\footnote{\url{https://github.com/ZubinGou/math-evaluation-harness}} for standardized eval.

\subsection{Knowledge QA with Search Retriever (VT-Search)}
\label{subsec:search_r1_details}

Question answering tasks often require access to external knowledge beyond the model's parametric memory, particularly for factual queries and multi-hop reasoning. We integrate a FAISS-based search retriever tool that enables agents to query a local knowledge base and extract relevant information for answering complex questions.

Following prior work~\citep{searchr1,song2025r1searcherincentivizingsearchcapability}, we integrate an E5 retriever~\citep{Wang2022TextEB} and index the 2018 Wikipedia dump~\citep{Karpukhin2020DensePR}. The agent alternates between search operations and reasoning steps to construct comprehensive answers.

For this task, we apply accuracy as the primary reward:
\begin{equation}
    R_{\text{search}}(\mathbf{x},\mathbf{y}) = \begin{cases}
1 & \text{if } \text{match}(\mathbf{y}, \mathbf{y}_g) \\
-1 & \text{otherwise} 
\end{cases}
\end{equation}

We evaluate using Exact Match scores on General Q\&A benchmarks (NQ~\citep{kwiatkowski2019natural}, TriviaQA~\citep{joshi2017triviaqa}, PopQA~\citep{mallen2022not}) and Multi-hop Q\&A benchmarks (HotpotQA~\citep{yang2018hotpotqa}, 2Wiki~\citep{ho2020constructing}, MuSiQue~\citep{trivedi2022musique}, Bamboogle~\citep{press2022measuring}).

\subsection{Multi-Turn SQL Query Generation (VT-SQL)}
\label{subsec:sql_executor_details}

Natural language-to-SQL (NL2SQL) conversion requires understanding database schemas and translating natural language queries into executable SQL commands. This task benefits from tool integration as it allows iterative query refinement based on execution feedback and error correction.

We assess SQL Executor adaptability using the SkyRL-SQL training set~\citep{liu2025skyrlsql} with Qwen2.5-7B-Instruct as the base model. The agent translates natural language questions into executable SQL, given schema hints and tool-calling instructions.

The reward function focuses solely on execution accuracy:
\begin{equation}
    R_{\text{sql}}(\mathbf{x},\mathbf{y}) = \begin{cases}
1 & \text{if } \text{match}(\mathbf{y}, \mathbf{y}_g) \\
-1 & \text{otherwise} 
\end{cases}
\end{equation}

Following standard conventions, we evaluate execution accuracy (EX) on \textsc{Spider-1.0}~\citep{Yu2018SpiderAL} (Dev and Test splits), \textsc{Spider-DK}~\citep{gan2021exploring}, and \textsc{Spider-Syn}~\citep{gan-etal-2021-towards}.

\subsection{Visual Reasoning with Image Operations (VT-VisualReasoner)}
\label{subsec:pixel_reasoner_details}

Traditional visual reasoning tasks are conducted primarily in the text modality, where models cannot dynamically process images as actions. To address this limitation, we implement image operation tools that enable agents to zoom into specific image regions, select key frames, and perform other visual manipulations to enhance reasoning over dense visual information, following the Pixel-Reasoner~\citep{pixelreasoner} approach.

The reward design incorporates both accuracy-oriented and compositional complexity measures:
\begin{alignat}{2}
R_{\text{visualreasoner}}(\mathbf{x},\mathbf{y}) &= r(\mathbf{x},\mathbf{y}) + \alpha \cdot r_{\text{curiosity}}(\mathbf{x}, \mathbf{y}) + \beta \cdot r_{\text{penalty}}(\mathbf{y}),\label{eq:reward}\\
\text{where}\quad r_{\text{curiosity}}(\mathbf{x}, \mathbf{y}) &= \max(H - \text{RaPR}(\mathbf{x}), 0) \cdot \mathbf{1}_{\text{PR}}(\mathbf{y})\\
    r_{\text{penalty}}(\mathbf{y}) &= \min(N - \mathbf{n}_{\text{vo}}(\mathbf{y}), 0)
\end{alignat}
where $\text{RaPR}(\mathbf{x})$ denotes the ratio of responses that invoke tool calls and $\mathbf{n}_{\text{vo}}(\mathbf{y})$ denotes the number of actions that response $\mathbf{y}$ invokes. Hyperparameters are set as $H=0.3$, $N=1$, $\alpha=0.5$ and $\beta=0.05$. We train two variants using accuracy-only reward and the original complexity-driven reward, denoted as ``GRPO-acc'' and ``GRPO-complex'' respectively.

We use the official training dataset from Pixel-Reasoner and evaluate primarily on V-Star~\citep{wu2024v}, which assesses MLLM visual search capabilities.

\subsection{Agentic Web Search (VT-DeeepSearch)}
\label{subsec:deepsearch_details}

Open-web question answering requires real-time information retrieval and multi-step reasoning over diverse web sources. GAIA~\citep{mialon2023gaia} and HLE~\citep{phan2025humanity} are representative benchmarks testing these capabilities. We implement a Web Search tool using Google Search API through SERPER with caching, enabling agents to perform dynamic information gathering and synthesis from online sources.

We apply both accuracy and tool-usage rewards to encourage effective search behavior:
\begin{equation}
R_{\text{deepsearch}}(\mathbf{x}, \mathbf{y}) = R_{\text{acc}}(\mathbf{x}, \mathbf{y}) + R_{\text{tool}}(\mathbf{x}, \mathbf{y}), \quad \text{where } R_{\text{tool}}(\mathbf{x}, \mathbf{y}) = \begin{cases}
    0.1, & \text{if tool is called} \\
    0, & \text{if no tool call}
\end{cases}
\end{equation}

We use 1K mixed training examples from SimpleDeepSearcher~\citep{sun2025simpledeepsearcher} and Web-Sailor~\citep{Li2025WebSailorNS}, following the setting in~\citet{dong2025agentic}. Starting from Qwen3-8B, we evaluate on GAIA and HLE (text-only) benchmarks. We retrieve top-k URLs for each query and use the returned snippets as content during RL training. For evaluation, we employ two settings: ``Snippet-Only'' aligns with training conditions using only snippet content, while ``QwQ-32B'' uses a browser agent to summarize raw content from retrieved URLs.

\subsection{Software Engineering Benchmark (VT-SWE)}
\label{subsec:swe_details}

Software engineering tasks require code understanding, localization, debugging, and modification capabilities that benefit from iterative execution and testing. We integrate bash terminal and code execution tools to enable agents to interact with software development environments effectively.

We build on the R2E-Gym scaffold~\citep{Jain2025R2EGymPE} and its training dataset R2E-Lite, using Qwen3-8B in no-think mode as the base model. The reward function is defined strictly by task completion accuracy: an agent must terminate normally and pass all verification tests to receive a reward of 1; otherwise, the reward is 0:
\begin{equation}
R_{\text{swe}}(\mathbf{x},\mathbf{y}) = \begin{cases}
1 & \text{if execution terminates successfully and all tests pass} \\
0 & \text{otherwise}
\end{cases}
\end{equation}

We evaluate performance on the SWE-Verified benchmark, measuring the agent's ability to resolve software engineering tasks and pass verification tests.

For training and evaluation, we maintain a cluster of eight servers (each with 64 CPU cores and 200 GB memory), orchestrating sandbox services via Kubernetes. Each task runs inside the official Docker image provided by R2E-Lite, initialized with 1 CPU and 2 GB memory, elastically scalable up to 2 CPUs and 4 GB memory. We observed that the main bottleneck lies in disk I/O during Docker initialization. To stabilize training, we therefore allocate more CPU and memory resources than are minimally required.

The modular architecture of \ours{}, which separates training from environment services, allows us to scale sandbox environments efficiently. Each environment interaction is given a 90-second timeout, reward computation has a 300-second timeout, and the maximum time per trajectory is capped at 20 minutes. Any trajectory that times out, encounters an exception, or exceeds the length limit is assigned a reward of 0, and its gradients are masked during updates.

\subsection{Supported Tools}
\label{subsec:supported_tools}







\begin{table}[!ht]
\centering
\small
\caption{Currently supported tools of \ours{} framework.}
\label{tab:all_tools}
\begin{tabular}{l|p{5.2cm}|p{4.7cm}}
\toprule
\textbf{Tools} & \textbf{Description} & \textbf{Related works} \\
\midrule
\textcolor{pythonblue}{\faPython} \textbf{Code Interpreter} & 
Execute Python code & 
ToRL~\citep{Li2025ToRLST} \\
\midrule
\textcolor{searchgreen}{\faSearch} \textbf{Faiss Search} & 
Vector similarity search for documents & 
Search-R1~\citep{searchr1} \\
\midrule
\textcolor{webblue}{\faGlobe} \textbf{Web Search API} & 
Real-time web search and retrieval & 
SimpleDeepSearch~\citep{sun2025simpledeepsearcher} \\
\midrule
\textcolor{imagepurple}{\faImage} \textbf{Image Processing} & 
Image resize, video frame selection & 
PixelReasoner~\citep{pixelreasoner} \\
\midrule
\textcolor{terminalorange}{\faTerminal} \textbf{Bash Terminal} & 
Execute shell commands & 
R2E-Gym~\citep{Jain2025R2EGymPE} \\
\midrule
\textcolor{sqlred}{\faDatabase} \textbf{SQL Executor} & 
Database queries and data management & 
SkySQL~\citep{liu2025skyrlsql} \\
\midrule
\textcolor{mcpgray}{\faPlug} \textbf{MCP Interface} & 
Model Context Protocol for external tool & 
ToolRL~\citep{Feng2025ReToolRL} \\
\bottomrule
\end{tabular}
\end{table}

\subsection{Training and Evaluation Configurations}
\label{subsec:training_configs}

\autoref{tab:training_params} summarizes the detailed configurations for each task during training and evaluation. Due to configuration differences across tasks, there may be gaps between validation curves and final downstream evaluation performance, as illustrated in Figure~\ref{fig:train_dynamics}.

The RL training parameters vary across tasks to accommodate different complexity levels and interaction patterns. Math-TIR and Pixel-Reasoner use smaller batch sizes due to computational constraints, while Search-R1 employs larger batch sizes for stable retrieval learning. The agentic tool use parameters reflect task-specific requirements: Math-TIR typically requires single-turn interactions, while SWE tasks may require up to 100 interaction turns for complex debugging scenarios.

Evaluation parameters are configured to balance comprehensive assessment with computational efficiency. Temperature settings range from 0.0 for deterministic tasks like SQL generation to 0.6 for creative tasks requiring exploration. Maximum turn limits reflect task complexity, with simple QA tasks limited to 2-5 turns while software engineering tasks allow up to 100 turns for thorough problem resolution.

\begin{table}[!bh]
\centering
\scriptsize
\setlength{\tabcolsep}{4pt}
\renewcommand{\arraystretch}{1.15}
\caption{Training and evaluation configurations across all six tasks.}
\label{tab:training_params}
\begin{tabular}{lcccccc}
\toprule
\textbf{Tasks} & \textbf{VT-Math} & \textbf{VT-Search} & \textbf{VT-SQL} & \textbf{VT-VisualReasoner} & \textbf{VT-DeepSearch} & \textbf{VT-SWE} \\
\midrule
\multicolumn{7}{c}{\textbf{RL Training Parameters}} \\
\midrule
\texttt{Rollout BS} & 128 & 512 & 256 & 128 & 128 & 32 \\
\texttt{N Samples} & 16 & 16 & 5 & 8 & 16 & 8 \\
\texttt{Gradient BS} & 128 & 64 & 256 & 128 & 128 & 32 \\
\texttt{Temperature} & 1.0 & 1.0 & 0.6 & 1.0 & 1.0 & 1.0 \\
\texttt{Top P} & 1.0 & 1.0 & 0.95 & 1.0 & 1.0 & 1.0 \\
\texttt{Learning Rate} & 1e-6 & 1e-6 & 1e-6 & 1e-6 & 1e-6 & 2e-6 \\
\texttt{Val Temperature} & 0.0 & 0.0 & 0.0 & 0.0 & 0.0 & 0.0 \\
\texttt{Val Top P} & 1.0 & 1.0 & 0.95 & 1.0 & 1.0 & 1.0 \\
\midrule
\multicolumn{7}{c}{\textbf{Agentic Tool Use Parameters}} \\
\midrule
\texttt{Max Turns} & 1 & 2 & 5 & 3 & 5 & 100 \\
\texttt{MTRL} & \ding{56} & \ding{56} & \ding{56} & \ding{52} & \ding{56} & \ding{52} \\
\texttt{Max Prompt Length} & 1024 & 4096 & 4096 & 16384 & 2048 & 10240 \\
\texttt{Max Response Length} & 3072 & 4096 & 4096 & 16384 & 8196 & 22528 \\
\texttt{Max Action Length} & 2048 & 2048 & 2048 & 2048 & 8196 & 10240 \\
\texttt{Max Observation Length} & 512 & 1024 & 1024 & 8192 & 4096 & 10240 \\
\multirow{2}{*}{\texttt{Action Stop Tokens}} & \multirow{2}{*}{\texttt{\texttt{`}\texttt{`}\texttt{`}output}} & \texttt{</search>,} & \multirow{2}{*}{\texttt{</sql>}} & \multirow{2}{*}{\texttt{</tool\_call>}} & \texttt{</python>,} & \multirow{2}{*}{\texttt{</function>}} \\
& & \texttt{</answer>} & & & \texttt{</search>} & \\
\midrule
\multicolumn{7}{c}{\textbf{Evaluation Parameters}} \\
\midrule
\texttt{Temperature} & 0.6 & 0.0 & 0.0 & 0.0 & 0.6 & 1.0 \\
\texttt{Top P} & 0.95 & 1.0 & 1.0 & 1.0 & 0.95 & 1.0 \\
\texttt{Max Turns} & 4 & 2 & 5 & 5 & 10 & 100 \\
\texttt{Max Prompt Length} & 1024 & 4096 & 4096 & 16384 & 2048 & - \\
\texttt{Max Response Length} & 3072 & 4096 & 4096 & 16384 & 32768 & 40960 \\
\texttt{Max Action Length} & 2048 & 2048 & 2048 & 4096 & 16483 & - \\
\texttt{Max Observation Length} & 512 & 1024 & 1024 & 8192 & 4096 & - \\
\bottomrule
\end{tabular}
\end{table}

\clearpage
\section{More related works}
\label{sec:more_related_works}

\subsection{Organization} 

In this section, we first establish the definition of Agentic Large Language Models (Agentic LLMs). Then, prior tool integration efforts in LLMs and the shift from single-turn, prompt-driven tool-calling to instruction tuning/RL-based multi-turn agentic interaction are reviewed. Further, we introduce various representative domain-specific tasks that are proven to benefit from developing corresponding tool-use oriented agents. We then distinguish Reinforcement Learning with Verifiable Rewards (RLVR) from Agentic Reinforcement Learning with Tool Use (ARLT). Finally, we survey existing systems for training RL-based tool-using agents and position our work: \ours{}.

\subsection{Tool-Integrated Reasoning}

Augmenting Large Language Models (LLMs) with external tools has become a prominent approach to address limitations in parametric reasoning and enable more complex task solving~\citep{Shen2024LLMWT}. Early approaches focused on \emph{prompt-based integration}, where systems like \textsc{HuggingGPT}~\citep{Shen2023HuggingGPTSA}, \textsc{Chameleon}~\citep{Lu2023ChameleonPC}, and \textsc{MultiTool-CoT}~\citep{Inaba2023MultiToolCoTGC} used tool manuals, demonstrations, or structured Chain-of-Thought templates to orchestrate sequential tool invocations. While these methods offer plug-and-play convenience, they face challenges when adapting to complex, multi-step tasks due to their reliance on static prompting strategies.

A complementary line of work explores \emph{instruction tuning}, where models are explicitly trained to recognize tool formats and generate appropriate function calls. Representative systems include \textsc{ToolFormer}~\citep{Schick2023ToolformerLM}, which uses bootstrapped annotations to teach tool usage patterns, \textsc{GPT4Tools}~\citep{Yang2023GPT4ToolsTL}, which distills tool-use traces from more capable models, and \textsc{LIMO}~\citep{Ye2025LIMOLI}, which demonstrates that targeted examples can elicit extended reasoning chains. However, these supervised approaches primarily provide static guidance and lack mechanisms for dynamic error correction based on tool execution feedback.

In contrast, reinforcement learning approaches enable models to develop adaptive tool-calling strategies through interaction-based training. Our work builds on this direction by employing GRPO training to enhance models' capabilities for self-reflection and iterative refinement in response to tool feedback.

\subsection{From Tool Integration to Agentic LLMs}
\label{subsec:agentic-def}

Large language models (LLMs) demonstrated exceptional flexibility and generality with model parameter and training-data scaling~\citep{gemini,qwen2025qwen25technicalreport,openai2024gpt41}. With recent research advancements dramatically enhancing their capability in reasoning, information retrieval, and instruction-following~\citep{Ke2025ASO}, current research trends have shifted from developing simple LLM-based tool-callers to empowering LLMs with versatile interaction capability to facilitate their agentic actions in the world~\citep{Plaat2025AgenticLL}. 

\paragraph{Agentic LLMs.}
As one of the central concepts in artificial intelligence~\citep{russell2016artificial}, Decision-making, identifying environmental changes, communication, and acting on one's goal or will are generally defined as common traits of Agency~\citep{epstein1996growing,wooldridge1999intelligent,gilbert2019agent}. Following well-established conventions, we denote Agentic LLMs as models that reason, act through tools, and interact over one or multiple turns, maintaining state and revising plans according to environmental observations, framing agent behavior beyond passive text-generation, and motivating the possession of improved planning and tool-use capabilities.

\paragraph{Agentic Tool-calling Acquisition.}

As one fundamental capability, tool-calling has been widely studied as one of the most effective ways of empowering Agentic LLMs with the capability to act in exposed environments. Early attempts involve prompt-based orchestration, which structures function calls in a training-free manner with specific instructions and tool-calling schemas~\citep{Yao2022ReActSR,Lu2023ChameleonPC}, 
exploiting LLMs' instruction-following capability through Chain-of-Thought technique~\citep{Wei2022ChainOT} and explicit task decomposition~\citep{Kim2023LanguageMC,Huang2022LanguageMA} or multi-agent orchestration~\citep{Shen2023HuggingGPTSA,Ruan2023TPTUTP}. Instruction-tuning based tool callers learn function-call schemas and appropriate tool-call choices through supervised traces~\citep{Schick2023ToolformerLM,Kong2023TPTUv2BT,gou2023tora}, improving problem-solving capabilities while suffering from a lack of generality and remaining largely single-turn~\citep{Qu2024ToolLW}. Reinforcement-Learning-based acquisition further enhances tool-using generality and multi-turn behaviors~\citep{Li2025ToRLST,Feng2025ReToolRL} through problem-solving outcomes and tool-feedback, facilitating environment exploration, self-reflection, error-corrections, and enhanced performance on reasoning-intensive tasks through informative tool responses.~\citep{Moshkov2025AIMO2WS}.

\subsection{Reinforcement Learning with Verifiable Reward (RLVR)}
\label{subsec:rlvr}

Conventional RLHF pipelines, such as DPO~\citep{Rafailov2023DirectPO} and PPO~\citep{Schulman2017ProximalPO}, optimize answer‐level quality and require reward models with substantial sizes. With the introduction of GRPO in \textsc{DeepSeekMath}~\citep{Shao2024DeepSeekMathPT}, as a critic-free variant of PPO, it enables stabilized RL training on long reasoning chains and enables the integration of multi-turn tool responses and verifiable rewards in reinforcement learning. Existing works have explored the potential of extending GRPO training through the inclusion of rule-based verifiable rewards such as format-based rewards and exact-match based comparisons, as well as tool-calling responses. They demonstrated significant success in developing expert agents in multiple domains with tool-calling and self-reflection capabilities. 


By integrating GRPO and verifiable rewards, advancements have been witnessed in a wide range of domain-specific tasks.  \textsc{ToRL}~\citep{Li2025ToRLST} integrates Python Code Interpreters into the GRPO training of solving mathematical tasks, surpassing RL baselines which does not have tool-calling integration on multiple mathematical datasets. \textsc{ToolRL}~\citep{Qian2025ToolRLRI} explored and analyzed the impact of reward design and tool choice on the effect of tool-integrated GRPO training, revealed the significance of tool-use reward design in boosting LLMs' tool calling and generalization capability, as well as achieving stable GRPO training. \textsc{ReTool}~\citep{Feng2025ReToolRL} integrated tool-calling in PPO training, resulting in a Python tool-enabled agent with strong mathematical problem-solving capability. \textsc{ExCoT}~\citep{Zhai2025ExCoTOR} and \textsc{Think2SQL}~\citep{Papicchio2025Think2SQLRL} demonstrated preliminary success of utilizing GRPO and the response of SQL executors to enhance the base model's performance on natural language-to-SQL (NL2SQL) tasks through comprehensive reward design and challenging training problem-filtering. By integrating Faiss~\citep{Douze2024TheFL}-based local retrievers and API-based online search services, \textsc{Search-R1}~\citep{searchr1} and \textsc{R1-Searcher}~\citep{song2025r1searcherincentivizingsearchcapability} equip base models with enhanced search tool-calling and retrieval capability, achieving superior performance across multiple retrieval-centric benchmarks.


\subsection{Agentic Reinforcement Learning with Tool Use (ARLT)}
\label{subsec:alrt}



Despite RLVR optimizing policies' behaviors using rule-based verifiable checks with the inclusion of single-turn tool-calling, current RLVR-based agentic models' limited capability in making real multi-turn, long-horizon interactions with tools and perform effective self-reflection and dynamic plan revising remains to be the gap to the generalist Agent-LLM. The inclusion of training the model over dynamic, long-term interactions while exposing it to intermediate tool responses could shape the agentic model's subsequent actions, pushing beyond single-turn verification toward long-horizon, interaction-centric RL training. 

Therefore, we use Agentic Reinforcement Learning with Tool Use (ARLT) to denote reinforcement learning on dynamic, multi-turn trajectories in which tool responses are treated as environmental observations that condition future actions. ARLT therefore (i) assigns credit across tool calls rather than only at the final answer, (ii) handles observation tokens explicitly (e.g., masking non-model tokens during GRPO optimization), and (iii) relies on asynchronous, failure-aware executors for adapting potentially slow, stochastic, or error-prone~\citep{Plaat2025AgenticLL,Ke2025ASO} tool calls. In contrast to RLVR, ARLT targets exploration, re-planning, and recovery from tool failures, and is naturally suited to settings where the problem-solving requires probing the environment before coming up with correct solutions. In this work, our framework is mainly evaluated on the following domain-specific tasks.

\paragraph{Mathematical Interactive Coding.}


Tool-integrated reasoning was first introduced to tackle computationally intensive mathematical problems by combining natural language reasoning with programming strategies~\citep{Chen2022ProgramOT,yue2023mammothbuildingmathgeneralist,searchr1,song2025r1searcherincentivizingsearchcapability,10.1145/3626772.3661381, Chen2022ProgramOT}. Building on this idea, \cite{wang2023mathcoderseamlesscodeintegration} proposed an iterative method that couples textual reasoning with code execution to cross-check the answers, improving the accuracy.
More recently, \cite{Slow_Thinking_with_LLMs_3_Tool} incorporated code execution into reasoning through supervised fine-tuning on curated code-integrated CoT data. Yet this method is limited by its dependence on specific data distributions and cannot learn adaptive tool-use strategies—such as when and how to call tools—via reinforcement learning. To solve this, concurrent work, including ToRL~\citep{Li2025ToRLST} and ZeroTIR~\citep{mai2025agent}, applies ZeroRL to train agents for mathematical code interpreter use.

\paragraph{Agentic Search and Retrieval.}
Large language models~\citep{geminiteam2024geminifamilyhighlycapable,openai2024gpt41} possess a huge amount of intrinsic knowledge while struggling at domain-specific, knowledge-centric tasks~\citep{Chen2024ASO,Peng2023ASO} and suffer from hallucination~\citep{Zhang2023SirensSI,Huang2023ASO}. A common approach to mitigate this issue is by integrating search engines into the LLMs. Predominant approaches of search engine integration often fall within two categories: Retrieval Augmented Generation (RAG)-based~\citep{Lewis2020RetrievalAugmentedGF,Gao2023RetrievalAugmentedGF} and Tool-calling based retrieval~\citep{Schick2023ToolformerLM}. As RAG relies on a separate retriever to extract documents in a single turn without the interaction of LLM, it faces challenges of retrieving irrelevant information or returning less useful context~\citep{Jin2024LongContextLM,Jiang2023ActiveRA}. Conversely, Tool-calling-based retrieval enhances LLMs' capability of calling the search retriever as a tool either through prompting~\citep{Yao2022ReActSR,Trivedi2022InterleavingRW}, fine-tuning~\citep{Schick2023ToolformerLM}, or through reinforcement learning~\citep{searchr1,song2025r1searcherincentivizingsearchcapability,Jiang2025DeepRetrievalHR}, while enhancing searching agents with multi-turn tool-calling-based retrieval capability remains under-explored.

\paragraph{Natural Language to SQL.}
Natural language-to-SQL (NL2SQL) refers to the task of generating database-specific query codes for extracting data of interest. Early efforts typically involve developing expert models with an encoder-decoder structure for achieving this goal, where the encoder fuses the database schema and the natural-language query, leaving the answer generation to the decoder module~\citep{Wang2019RATSQLRS, Cai2021SADGASD, Li2023GraphixT5MP,Liu2024ASO}. With the advances of sequence-to-sequence models, especially large language models, the NL2SQL tasks have seen a rapid transfer from simple sequence-to-sequence modeling into developing tabular understanding agents utilizing pre-trained large language models. Existing works mostly focus on prompt enhancement~\citep{Gao2023TexttoSQLEB}, large-scale data synthesis and model finetuning~\citep{Li2024CodeSTB,Li2025OmniSQLSH}, while some recent works explored the realm of Reinforcement Learning-driven NL2SQL agents~\citep{Papicchio2025Think2SQLRL,Pourreza2025ReasoningSQLRL} and NL2SQL agents featuring multi-round Tool Calling capabilities.~\citep{liu2025skyrlsql}

\clearpage
\section{Case Study}
\label{sec:case_study}

In this section, we present case studies of representative tasks, featuring corresponding agents' outputs to illustrate Multi-Turn Tool Calling. The queries are randomly extracted from the underlying benchmarks. For readability, tag markers are colorized according to their functional categories:

\begin{enumerate}
    \item Internal reasoning tags, \purple{\texttt{<think>...</think>}} are colored in purple.
    \item Tool-calling tags are formatted as \green{\texttt{<python>/<sql>/<search>}}.
    \item Observation/Tool Response tags are colored as \blue{\texttt{<observation>/<information>/<result>}}.
    \item The final output tags are formulated as \red{\texttt{<answer>/<solution>}}.
\end{enumerate}

We specifically chose the following four representative tasks for demonstration: \textbf{Mathematical Reasoning with Python Executor}, \textbf{Knowledge Q\&A with Search Retriever}, \textbf{Multi-turn SQL Generation}, and \textbf{Agentic Web Search}. For a detailed explanation of tag definitions, please refer to Tab.\ref{tab:training_params} and corresponding settings in Appendix's Sec. \ref{sec:exp_setup}. 

For Agentic Web Search, Multi-turn SQL Generation, and Knowledge Q\&A with Retriever, the text-heavy responses from the tools are truncated or simplified.

\begin{table*}[!ht]
    \centering
    \caption{Example answer demonstration of \textsc{VT-Math}}
    \fontsize{9pt}{11pt}\selectfont
    \begin{tabular}{p{0.98\linewidth}}
    \midrule
        \rowcolor{gray!20}\textbf{Example from DeepMath Dataset} \\
    \midrule
        \textbf{Question:} \\
        Determine the function $f$ given that it has a power series representation centered at $0$, converging within the interval $(-1, 1)$, and there exists a sequence of distinct points $(x_n)$ approaching $0$ such that $f(x_n) = 0$ for all $n \in \mathbb{N}$. What is $f$?\\
        \textbf{Labeled Answer:} \\
        $0$ \\
    \midrule
        \rowcolor{gray!20}
        \textbf{Model Output by VT-Math GRPO}\\
    \midrule
        Let's consider the problem. The function $f$ has a power series representation centered at $0$ that converges within the interval $(-1, 1)$. This means that $f(x)$ can be written as: $f(x) = \sum_{n=0}^{\infty} a_n x^n$ where the series converges for $|x| < 1$. We are given that there exists a sequence of distinct points $(x_n)$ approaching $0$ such that $f(x_n) = 0$ for all $n \in \mathbb{N}$. This implies that: $f(x_n) = \sum_{n=0}^{\infty} a_n x_n^n = 0$ Since the series converges at $x_n$ and $x_n \to 0$, we can conclude that each $a_n = 0$ for all $n \geq 1$ because if any $a_k \neq 0$ for some $k \geq 1$, then the series would not converge to $0$ as $x_n \to 0$. Therefore, the only possible term is the constant term $a_0$. This means we can write: \[ f(x) = a_0 \] Given that $f(0) = a_0 = 0$ (since $f(x_n) \to f(0) = 0$ as $x_n \to 0$), we have: \[ f(x) = 0 \] Therefore, the function $f$ is the zero function. Let's put this understanding into a more formal version:
        
        \green{\texttt{`}\texttt{`}\texttt{`}python}
        \begin{lstlisting}[style=python]
from sympy import symbols, Function, Eq, solve

# Define the variable and the function
x = symbols('x')
f = Function('f')(x)
# Define the power series representation centered at 0
# Since we are given that there exists a sequence (x_n) approaching 0 such that f(x_n) = 0 for all
# It implies that the function is the zero function
f_zero = 0
# Print the zero function
print(f_zero)
        \end{lstlisting}
        \green{\texttt{`}\texttt{`}\texttt{`}}

        \blue{\texttt{`}\texttt{`}\texttt{`}output} \\
        result \\
        0 \\
        \blue{\texttt{`}\texttt{`}\texttt{`}} \\

        The power series representation of the function $f$ converging within the interval $(-1, 1)$, given that there exists a sequence of distinct points $(x_n)$ approaching $0$ such that $f(x_n) = 0$ for all $n \in \mathbb{N}$, results in the function being the zero function. Therefore, the function $f$ is \boxed{\text{0}}.\\
        
    \bottomrule
    \end{tabular}
    \label{tab:case_math}
\end{table*}
\begin{table*}[!t]
    \centering
    \caption{Example answer demonstration of \textsc{VT-Search}}
    \fontsize{9pt}{11pt}\selectfont
    \begin{tabular}{p{0.98\linewidth}}
    \midrule
        \rowcolor{gray!20}\textbf{Example from HotpotQA Dataset} \\
    \midrule
        \textbf{Question:} \\
        What is the birth name of Nadeem Siddique's favorite boxer?\\
        \textbf{Labeled Answer:} \\
        Walker Smith Jr. \\
    \midrule
        \rowcolor{gray!20}
        \textbf{Model Output by VT-Search (GRPO)}\\
    \midrule

        \purple{<think>} \\
        
        To answer this question, I need to find out Nadeem Siddique's favorite boxer and then determine the birth name of that boxer. \\
        
        \purple{</think>} \\

        \green{<search>}\\
        Who is Nadeem Siddique's favorite boxer? \\
        \green{</search>} \\

        \blue{<information>}\\ 
        
        Doc 1(Title: "Nadeem Siddique") Nadeem Siddique Siddique was born in Bradford, Yorkshire, United Kingdom. He supports his local football team Bradford City and \textbf{his favorite boxer is Sugar Ray Robinson}... \\
        Doc 2(Title: "Nadeem Siddique") in 2005 for the people of Palestine and Gaza whereby he raised £352,000. He stated 'Me and Junior have been doing loads of things for charities and with our popularity in Bradford we always raise quite a few quid...  \\
        Doc 3(Title: "Nadeem Siddique") 3 fights and 3 wins and is on the verge of a major championship fight. Siddique is the founder of 'Make the Weight' gym; the opening of the gym was attended by boxer Mike Tyson in whose... \\
        
        \blue{</information>} \\

        From the search results, I found that Nadeem Siddique's favorite boxer is Sugar Ray Robinson. \\

        \green{<search>}\\
        What is the birth name of Sugar Ray Robinson? \\
        \green{</search>} \\

        \blue{<information>} \\
        Doc 1(Title: "Sugar Ray Robinson") Robinson was born \textbf{Walker Smith Jr}. in Ailey, Georgia, to Walker Smith Sr. and Leila Hurst. Robinson was the youngest of three children...  \\
        Doc 2(Title: "Sugar Ray Robinson") disease. He died in Los Angeles at the age of 67 and was interred in the Inglewood Park Cemetery, Inglewood, California. Robinson married Marjorie Joseph in 1938; the marriage was annulled the same year. Their son, Ronnie Smith... \\
        Doc 3(Title: "Sugar Ray Robinson") Jackie Gleason, Nat King Cole, Joe Louis, and Lena Horne among others. Robinson was known as a flamboyant personality outside the ring. He combined striking good looks with charisma and a flair for the dramatic... \\
        \blue{</information>} \\

        From the search results, I found that the birth name of Sugar Ray Robinson is Walker Smith Jr. \\

        \red{<answer>} Walker Smith Jr. \red{</answer>} \\

    \bottomrule
    \end{tabular}
    \label{tab:case_searchr1}
\end{table*}

\begin{table*}[!t]
    \centering
    \caption{Example answer demonstration (SQL reasoning)}
    \fontsize{9pt}{11pt}\selectfont
    \begin{tabular}{p{0.98\linewidth}}
    \midrule
        \rowcolor{gray!20}\textbf{Example (Student--Pet DB)} \\
    \midrule
        \textbf{Question:} \\
        What are the ids of the students who do not own cats as pets? \\
        
        \textbf{Labeled Answer:} \\
        \begin{lstlisting}[style=sql]
SELECT stuid FROM student EXCEPT SELECT T1.stuid FROM student AS T1 JOIN has_pet AS T2 ON T1.stuid  =  T2.stuid JOIN pets AS T3 ON T3.petid  =  T2.petid WHERE T3.pettype  =  'cat'
        \end{lstlisting} \\
        
    \midrule
        \rowcolor{gray!20}
        \textbf{Model Output (Agentic SQL)}\\
    \midrule

\purple{<think>}I need to find the student IDs of students who do not own cats as pets. To achieve this, I will first identify all students who own cats and then exclude them from the list of all students.\\
\purple{</think>} \\

\green{<sql>} \\

\begin{lstlisting}[style=sql]
SELECT StuID FROM Has_Pet WHERE PetID IN (SELECT PetID FROM Pets WHERE PetType = 'cat');

\end{lstlisting}

\green{</sql>} \\

\blue{<observation>} \\
0 \\
1001 \\
<reminder>You have 5 turns left to complete the task.</reminder> \\

\blue{</observation>} \\

\purple{<think>}I have the list of student IDs who own cats. Now, I will use a NOT IN clause to exclude these IDs from the list of all students. \\
\purple{</think>} \\

\green{<sql>}
\begin{lstlisting}[style=sql]
SELECT StuID FROM Student WHERE StuID NOT IN (SELECT StuID FROM Has_Pet WHERE PetID IN (SELECT PetID FROM Pets WHERE PetType = 'cat'));

\end{lstlisting}
\green{</sql>} \\

\blue{<observation>} \\
0 \\
1002 \\
1003 \\
... \\

<reminder>You have 4 turns left to complete the task.</reminder> \\

\blue{</observation>} \\

\purple{<think>}I have the list of student IDs who do not own cats as pets. Since the question asks for the ids of the students, I can now output the final SQL as the solution. \\
\purple{</think>} \\

\red{<solution>} \\

\begin{lstlisting}[style=sql]
SELECT StuID FROM Student WHERE StuID NOT IN (SELECT StuID FROM Has_Pet WHERE PetID IN (SELECT PetID FROM Pets WHERE PetType = 'cat'));

\end{lstlisting}

\red{</solution>} \\

    \bottomrule
    \end{tabular}
    \label{tab:case_sql}
\end{table*}

\begin{table*}[!t]
    \centering
    \caption{Example answer demonstration of \textsc{VT-DeepSearch}}
    \fontsize{9pt}{11pt}\selectfont
    \begin{tabular}{p{0.98\linewidth}}
    \midrule
        \rowcolor{gray!20}\textbf{Example from Gaia Dataset} \\
    \midrule
        \textbf{Question:} \\
        Find the value of $x$ to the nearest tenth: Lx = (d/dx * (A * x-squared)) + 4-thousand'n'ninety-7 minus C 
        
        Where L is the last two digits of the year of the Venezuelan Declaration of Independence, A is the number of colors in the TikTok logo as of July 2023, excluding black and white, and C is the height of the average woman in the Philippines according to a July 2023 Business Insider article, rounded to the nearest whole centimeter. \\
        \textbf{Labeled Answer:} \\
        563.9 \\
    \midrule
        \rowcolor{gray!20}
        \textbf{Model Output by VT-DeepSearcher (GRPO)}\\
    \midrule

        \purple{<think>} Okay, let's try to figure out this problem step by step. First, I need to understand what each part of the equation means and find the values for L, A, and C. Then I can plug them into the equation to solve for x.

        The equation given is Lx = (d/dx * (A * x-squared)) + 4-thousand'n'ninety-7 minus C. Wait, the wording here is a bit confusing. Let me parse it again. Maybe it's L multiplied by x equals the derivative of (A times x squared) plus 4097 minus C. Or maybe there's a typo? Let me check the original question again.

        The equation is written as: Lx = (d/dx * (A * x-squared)) + 4-thousand'n'ninety-7 minus C. Hmm, the "4-thousand'n'ninety-7" part might be a way to write 4097. So maybe it's 4097. So the equation is Lx equals the derivative of A x squared plus 4097 minus C. \\

        \green{<search>} Venezuelan Declaration of Independence year \green{</search>} \\
        \blue{<result>} \\
        The Venezuelan Declaration of Independence occurred on \textbf{**July 5, 1811**}. The last two digits of this year are \textbf{**11**}, which corresponds to L...\\
        \blue{</result>} \\
        So L = 11. \\
        \green{<search>} TikTok logo colors 2023 \green{</search>} \\
        \blue{<result>} The TikTok logo as of 2023 features \textbf{**two colors excluding black and white**} : pink (\#FE2C55) and aqua/cyan (\#25F4EE). Thus A = 2... \\
        \blue{</result>} \\
        So A = 2. \\

        \green{<search>} average height of women in Philippines 2023 Business Insider \green{</search>} \\
        \blue{<result>} The average height of women in the Philippines is \textbf{**149.6 cm**}, rounded to \textbf{**150 cm**}. Thus C = 150...\\
        \blue{</result>} \\
        So C = 150. \\

        Now the equation is: \\
        $Lx = (d/dx (A x^2)) + 4097 - C$. \\
        Derivative of $A x^2$ is $2Ax = 4x$. \\
        Substituting $L=11, A=2, C=150$: \\
        $11x = 4x + 4097 - 150 = 4x + 3947$. \\
        Simplify: $7x = 3947 \Rightarrow x = 3947/7$. \\

        \blue{<python>}
        \begin{lstlisting}[style=python]
3947 / 7
        \end{lstlisting}
        \blue{</python>} \\
        \blue{<result>} \\
        563.9 \\
        \blue{</result>} \\
        \red{<answer>} \\
        The final answer is $\boxed{563.9}$. \\
        \red{</answer>} \\
        \purple{</think>} \\
    \bottomrule
    \end{tabular}
    \label{tab:case_deepsearch}
\end{table*}

\clearpage
\section{The use of Large Language Models}
We have utilized Claude and ChatGPT's website clients primarily for writing enhancement tasks, including eliminating potential grammar errors, improving the logical coherence of existing content, and reorganizing tables and figures for better presentation. Additionally, we employed these tools for data visualization, such as generating ~\autoref{fig:train_dynamics} from our collected data. We also leveraged ChatGPT's deep research capabilities to identify relevant works in agentic RL and infrastructure, which are comprehensively presented in ~\autoref{sec:related} and ~\autoref{sec:more_related_works}.


\end{document}